\documentclass{article}
\usepackage{color}
\usepackage{amssymb,amsmath,graphicx}
\usepackage[usenames,dvipsnames,svgnames,table]{xcolor}
\usepackage{amsfonts}
\usepackage{url}
\usepackage[capitalise]{cleveref}

\newcounter{mnote}

\newcommand{\ie}{i.e.\ }
\newcommand{\eg}{e.g.\ }
\newcommand{\cf}{c.f.\ }

\newcommand{\flabel}[1]{\label{fig:#1}}
\newcommand{\seclabel}[1]{\label{sec:#1}}
\newcommand{\tlabel}[1]{\label{tab:#1}}

\newcommand{\fref}[1]{Fig.~\ref{fig:#1}}
\newcommand{\sref}[1]{Section~\ref{sec:#1}}
\newcommand{\tref}[1]{Table~\ref{tab:#1}}

\newcommand*\idx[2][]
{ 
\def\next{#1}%
\ifx\empty\next
  (#2)
\else
  (#1, #2)
\fi
}
\newcommand*\elt[3][]
{ 
\def\next{#1}%
\ifx\empty\next
  #2\idx{#3}
\else
  #1\idx{#2,#3}
\fi
}
\newcommand*\pd[3][]
{ 
\def\next{#1}%
\ifx\empty\next
  \frac{\partial#2}{\partial #3}
\else
  \frac{{\partial^{#1} #2}}{\partial#3^{#1}}
\fi
}

\newcommand{\igate}{i}
\newcommand{\fgate}{f}
\newcommand{\ogate}{o}
\newcommand{\state}{c}
\newcommand{\hiddenfn}{\mathcal{H}}
\newcommand{\outputfn}{\mathcal{Y}}

\newcommand{\wtmat}[2]{W_{#1 #2}}

\newcommand{\bias}[1]{b_{#1}}
\newcommand{\hbias}{\bias{h}}
\newcommand{\obias}{\bias{y}}
\newcommand{\seq}[1]{\mathbf{#1}}

\newcommand{\invble}{x}
\newcommand{\outvble}{y}

\newcommand{\inseq}{\seq{\invble}}
\newcommand{\ann}{\seq{c}}
\newcommand{\outseq}{\seq{\outvble}}

\newcommand{\loss}{\mathcal{L}(\inseq)}
\newcommand{\aloss}{\mathcal{L}(\inseq)}

\newcommand{\reals}{\mathbb{R}}

\newcommand{\expo}[1]{\exp\left(#1\right)}

\newcommand{\gauss}{\mathcal{N}}

\newcommand{\deq}{\mathrel{\stackrel{\text{\tiny{def}}}{=}}}

\newcommand{\pred}{\Pr(x_{t+1}|y_t)}

\newcommand{\figdir}{}
\newcommand{\capt}[2]{\caption[#1]{\textbf{#1}#2}}

\newcommand{\fig}[5]
{
\begin{figure}
\begin{center}
\includegraphics[width=#3\columnwidth]{\figdir#1}
\end{center}
\capt{#4}{#5}
\flabel{#2}
\end{figure}
}

\begin{document}
\title{Generating Sequences With\\Recurrent Neural Networks}
\author{Alex Graves\\Department of Computer Science\\University of Toronto\\ \texttt{graves@cs.toronto.edu}}
\date{}
\maketitle
\begin{abstract}
This paper shows how Long Short-term Memory recurrent neural networks can be used to generate complex sequences with long-range structure, simply by predicting one data point at a time. 
The approach is demonstrated for text (where the data are discrete) and online handwriting (where the data are real-valued). 
It is then extended to handwriting synthesis by allowing the network to condition its predictions on a text sequence. 
The resulting system is able to generate highly realistic cursive handwriting in a wide variety of styles.
\end{abstract}

\section{Introduction}
%
%

Recurrent neural networks (RNNs) are a rich class of dynamic models that have been used to generate sequences in domains as diverse as music~\cite{eck02music,boulanger12music}, text~\cite{sutskever11rnn} and motion capture data~\cite{sutskever08rtbm}.
RNNs can be trained for sequence generation by processing real data sequences one step at a time and  predicting what comes next.
Assuming the predictions are probabilistic, novel sequences can be generated from a trained network by iteratively sampling from the network's output distribution, then feeding in the sample as input at the next step. 
In other words by making the network treat its inventions as if they were real, much like a person dreaming.
Although the network itself is deterministic, the stochasticity injected by picking samples induces a distribution over sequences.
This distribution is conditional, since the internal state of the network, and hence its predictive distribution, depends on the previous inputs.



RNNs are `fuzzy' in the sense that they do not use exact templates from the training data to make predictions, but rather---like other neural networks---use their internal representation to perform a high-dimensional interpolation between training examples.
This distinguishes them from n-gram models and compression algorithms such as Prediction by Partial Matching~\cite{cleary84ppm}, whose predictive distributions are determined by counting exact matches between the recent history and the training set.
The result---which is immediately apparent from the samples in this paper---is that RNNs (unlike template-based algorithms) synthesise and reconstitute the training data in a complex way, and rarely generate the same thing twice.
Furthermore, fuzzy predictions do not suffer from the curse of dimensionality, and are therefore much better at modelling real-valued or multivariate data than exact matches.


In principle a large enough RNN should be sufficient to generate sequences of arbitrary complexity.
In practice however, standard RNNs are unable to store information about past inputs for very long~\cite{hochreiter01book}.
As well as diminishing their ability to model long-range structure, this `amnesia' makes them prone to instability when generating sequences.
The problem (common to all conditional generative models) is that if the network's predictions are only based on the last few inputs, and these inputs were themselves predicted by the network, it has little opportunity to recover from past mistakes.
Having a longer memory has a stabilising effect, because even if the network cannot make sense of its recent history, it can look further back in the past to formulate its predictions.
The problem of instability is especially acute with real-valued data, where it is easy for the predictions to stray from the manifold on which the training data lies.
One remedy that has been proposed for conditional models is to inject noise into the predictions before feeding them back into the model~\cite{taylor09gait}, thereby increasing the model's robustness to surprising inputs.
However we believe that a better memory is a more profound and effective solution.

Long Short-term Memory (LSTM)~\cite{hochreiter97lstm} is an RNN architecture designed to be better at storing and accessing information than standard RNNs.
LSTM has recently given state-of-the-art results in a variety of sequence processing tasks, including speech and handwriting recognition~\cite{graves13icassp,graves09nips}.
The main goal of this paper is to demonstrate that LSTM can use its memory to generate complex, realistic sequences containing long-range structure.

\sref{pred_net} defines a `deep' RNN composed of stacked LSTM layers, and explains how it can be trained for next-step prediction and hence sequence generation.
\sref{text} applies the prediction network to text from the Penn Treebank and Hutter Prize Wikipedia datasets.
The network's performance is competitive with state-of-the-art language models, and it works almost as well when predicting one character at a time as when predicting one word at a time.
The highlight of the section is a generated sample of Wikipedia text, which showcases the network's ability to model long-range dependencies.
\sref{hand_pred} demonstrates how the prediction network can be applied to real-valued data through the use of a mixture density output layer, and provides experimental results on the IAM Online Handwriting Database.
It also presents generated handwriting samples proving the network's ability to learn letters and short words direct from pen traces, and to model global features of handwriting style.
\sref{hand_synth} introduces an extension to the prediction network that allows it to condition its outputs on a short annotation sequence whose alignment with the predictions is unknown.
This makes it suitable for handwriting synthesis, where a human user inputs a text and the algorithm generates a handwritten version of it.
The synthesis network is trained on the IAM database, then used to generate cursive handwriting samples, some of which cannot be distinguished from real data by the naked eye.
A method for biasing the samples towards higher probability (and greater legibility) is described, along with a technique for `priming' the samples on real data and thereby mimicking a particular writer's style.
Finally, concluding remarks and directions for future work are given in \sref{conclusion}.

%
%

\section{Prediction Network}
\seclabel{pred_net}
\fref{deep_predictor} illustrates the basic recurrent neural network prediction architecture used in this paper.
An input vector sequence $\inseq = (x_1,\ldots,x_T)$ is passed through weighted connections to a stack of $N$ recurrently connected hidden layers to compute first the hidden vector sequences $\seq{h}^n = (h^n_1,\ldots,h^n_T)$ and then the output vector sequence $\outseq = (y_1,\ldots,y_T)$.
Each output vector $y_t$ is used to parameterise a predictive distribution $\Pr(x_{t+1}|y_t)$ over the possible next inputs $x_{t+1}$.
The first element $x_1$ of every input sequence is always a null vector whose entries are all zero; the network therefore emits a prediction for $x_2$, the first real input, with no prior information.
The network is `deep' in both space and time, in the sense that every piece of information passing either vertically or horizontally through the computation graph will be acted on by multiple successive weight matrices and nonlinearities.

\fig{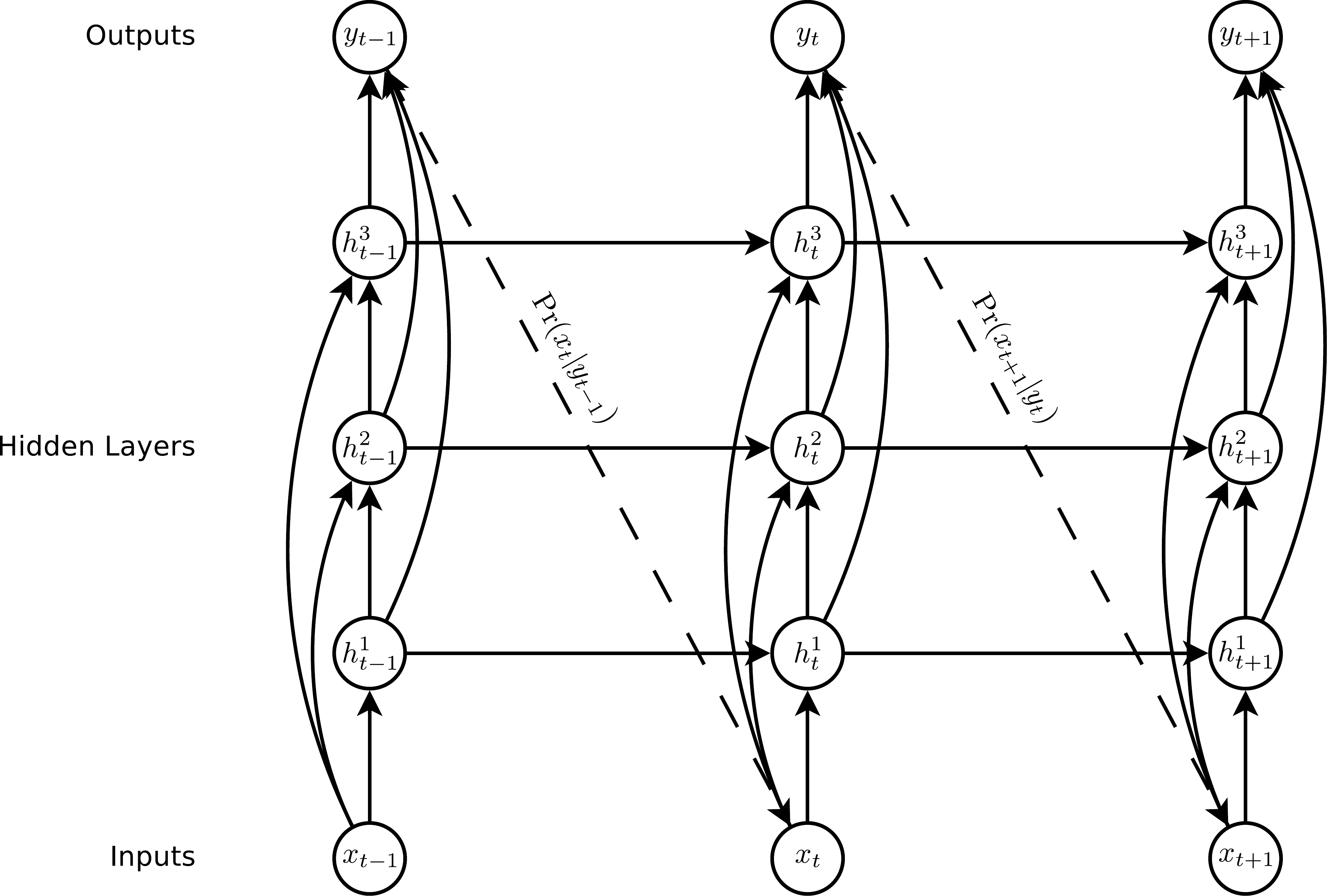}{deep_predictor}{0.925}{Deep recurrent neural network prediction architecture.}{ The circles represent network layers, the solid lines represent weighted connections and the dashed lines represent predictions.}

Note the `skip connections' from the inputs to all hidden layers, and from all hidden layers to the outputs.
These make it easier to train deep networks, by reducing the number of processing steps between the bottom of the network and the top, and thereby mitigating the `vanishing gradient' problem~\cite{bengio94learning}.
In the special case that $N=1$ the architecture reduces to an ordinary, single layer next step prediction RNN.

The hidden layer activations are computed by iterating the following equations from $t=1$ to $T$ and from $n=2$ to $N$:
\begin{align}
\label{eq:pred_hidden}
h^1_t &= \hiddenfn\left(W_{i h^1} x_t + W_{h^{1}h^{1}} h^1_{t-1} + \hbias^1 \right)\\
h^n_t &= \hiddenfn\left(W_{i h^n} x_t + W_{h^{n-1}h^{n}} h^{n-1}_t + W_{h^{n}h^{n}} h^n_{t-1} + \hbias^n \right)
\end{align} 
where the $W$ terms denote weight matrices (\eg $W_{i h^n}$ is the weight matrix connecting the inputs to the $n^{th}$ hidden layer, $W_{h^{1}h^{1}}$ is the recurrent connection at the first hidden layer, and so on), the $b$ terms denote bias vectors (\eg $\obias$ is output bias vector) and $\hiddenfn$ is the hidden layer function. 

Given the hidden sequences, the output sequence is computed as follows:
\begin{align}
\label{eq:pred_output}
\hat{y}_t &= \obias + \sum_{n=1}^N{W_{h^n y} h^n_t}\\
y_t &= \outputfn(\hat{y}_t)	
\end{align}
where $\outputfn$ is the output layer function.
The complete network therefore defines a function, parameterised by the weight matrices, from input histories $\inseq_{1:t}$ to output vectors $y_t$.

The output vectors $y_t$ are used to parameterise the predictive distribution $\Pr(x_{t+1}|y_t)$ for the next input.
The form of $\Pr(x_{t+1}|y_t)$ must be chosen carefully to match the input data.
In particular, finding a good predictive distribution for high-dimensional, real-valued data (usually referred to as \emph{density modelling}), can be very challenging.

The probability given by the network to the input sequence $\inseq$ is
\begin{equation}
\Pr(\inseq) = \prod_{t=1}^T{\Pr(x_{t+1}|y_t)}
\end{equation}
and the sequence loss $\loss$ used to train the network is the negative logarithm of $\Pr(\inseq)$:
\begin{equation}
\label{eq:loss}
\loss = -\sum_{t=1}^T{\log \Pr(x_{t+1}|y_t)}
\end{equation}
The partial derivatives of the loss with respect to the network weights can be efficiently calculated with backpropagation through time~\cite{williams95bptt} applied to the computation graph shown in \fref{deep_predictor}, and 
the network can then be trained with gradient descent.

\subsection{Long Short-Term Memory}
\seclabel{lstm}
In most RNNs the hidden layer function $\hiddenfn$ is an elementwise application of a sigmoid function.
However we have found that the Long Short-Term Memory (LSTM) architecture~\cite{hochreiter97lstm}, which uses purpose-built \emph{memory cells} to store information, is better at finding and exploiting long range dependencies in the data.
\fref{lstm} illustrates a single LSTM memory cell.
For the version of LSTM used in this paper~\cite{gers02peepholes} $\hiddenfn$ is implemented by the following composite function:
\begin{align}
\igate_t &= \sigma\left(\wtmat{x}{\igate} x_t + \wtmat{h}{\igate} h_{t-1} + \wtmat{\state}{\igate} \state_{t-1}  + b_\igate\right)\\
\fgate_t &= \sigma\left(\wtmat{x}{\fgate} x_t + \wtmat{h}{\fgate} h_{t-1} + \wtmat{\state}{\fgate} \state_{t-1} + b_\fgate \right)\\
\state_t &= \fgate_t \state_{t-1} + \igate_t \tanh \left(\wtmat{x}{\state} x_t + \wtmat{h}{\state} h_{t-1} + b_\state\right)\\
\ogate_t &= \sigma\left(\wtmat{x}{\ogate} x_t + \wtmat{h}{\ogate} h_{t-1} + \wtmat{\state}{\ogate} \state_{t} + b_\ogate\right)\\
h_t &= \ogate_t \tanh(\state_t)
\end{align}
where $\sigma$ is the logistic sigmoid function, and $\igate$, $\fgate$, $\ogate$ and $\state$ are respectively the \emph{input gate}, \emph{forget gate}, \emph{output gate}, \emph{cell} and \emph{cell input} activation vectors, all of which are the same size as the hidden vector $h$.
The weight matrix subscripts have the obvious meaning, for example $\wtmat{h}{\igate}$ is the hidden-input gate matrix, $\wtmat{x}{\ogate}$ is the input-output gate matrix etc.
The weight matrices from the cell to gate vectors (\eg $\wtmat{c}{\igate}$) are diagonal, so element $m$ in each gate vector only receives input from element $m$ of the cell vector.
The bias terms (which are added to $\igate$, $\fgate$, $\state$ and $\ogate$) have been omitted for clarity.

\fig{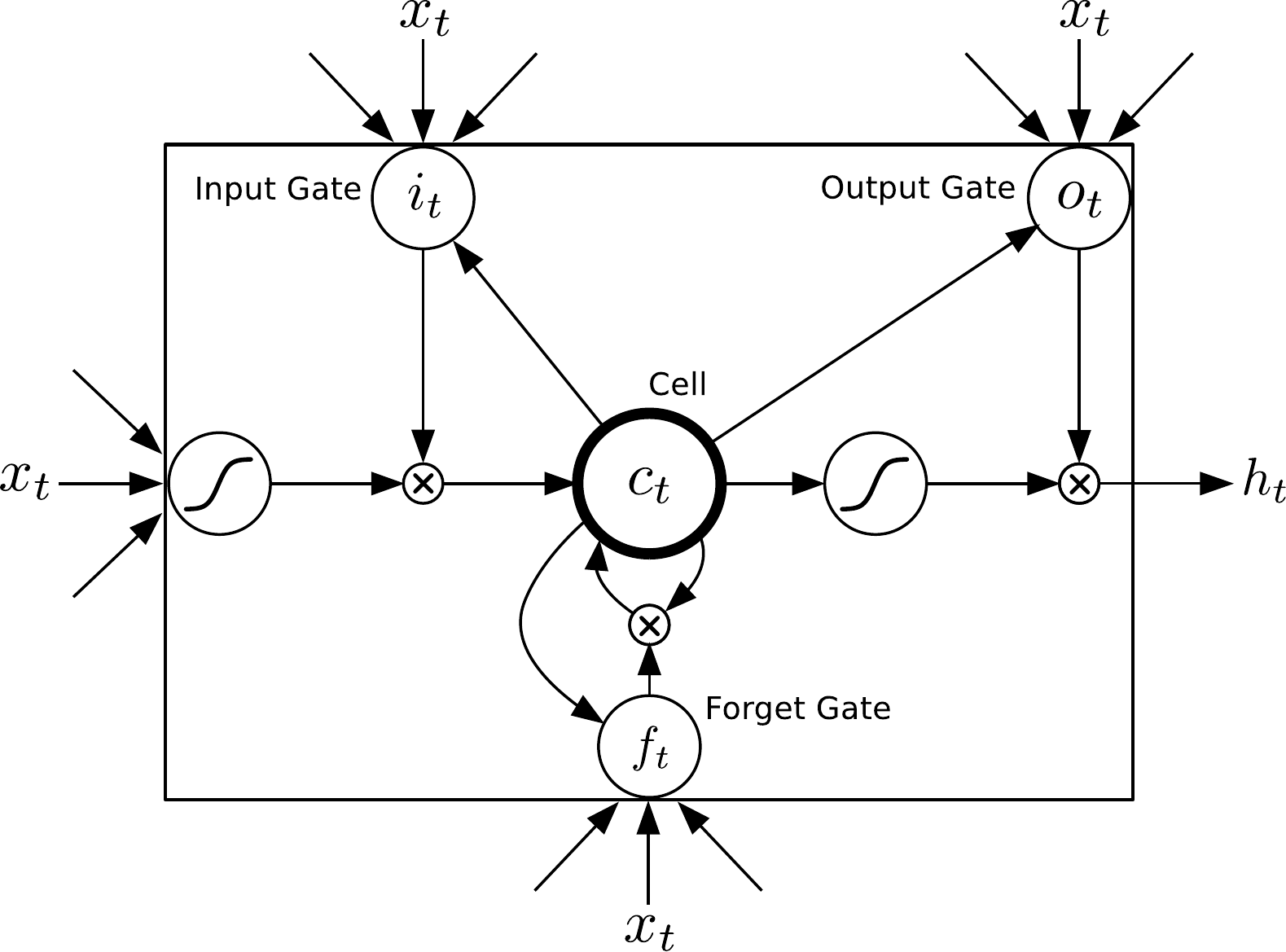}{lstm}{0.8}{Long Short-term Memory Cell}{}

The original LSTM algorithm used a custom designed approximate gradient calculation that allowed the weights to be updated after every timestep~\cite{hochreiter97lstm}.
However the full gradient can instead be calculated with backpropagation through time~\cite{graves05nn}, the method used in this paper.
One difficulty when training LSTM with the full gradient is that the derivatives sometimes become excessively large, leading to numerical problems.
To prevent this, all the experiments in this paper clipped the derivative of the loss with respect to the network inputs to the LSTM layers (before the sigmoid and $tanh$ functions are applied) to lie within a predefined range\footnote{In fact this technique was used in all my previous papers on LSTM, and in my publicly available LSTM code, but I forgot to mention it anywhere---\emph{mea culpa}.}.

\section{Text Prediction}
\seclabel{text}
Text data is discrete, and is typically presented to neural networks using `one-hot' input vectors.
That is, if there are $K$ text classes in total, and class $k$ is fed in at time $t$, then $x_t$ is a length $K$ vector whose entries are all zero except for the $k^{th}$, which is one.
$\pred$ is therefore a multinomial distribution, which can be naturally parameterised by a softmax function at the output layer:
\begin{equation}
\label{eq:softmax}
\Pr(x_{t+1}=k|y_t) = y^k_t = \frac{\expo{\hat{y}^k_t}}{\sum_{k'=1}^K{\expo{\hat{y}^{k'}_{t}}}}
\end{equation}
Substituting into Eq.~(6) we see that
\begin{align}
\loss &= -\sum_{t=1}^T{\log y^{x_{t+1}}_t}\\
\implies \pd{\loss}{\hat{y}^k_t} &= y^k_t - \delta_{k,x_{t+1}}	
\end{align}
The only thing that remains to be decided is which set of classes to use.
In most cases, text prediction (usually referred to as \emph{language modelling}) is performed at the word level.
$K$ is therefore the number of words in the dictionary.
This can be problematic for realistic tasks, where the number of words (including variant conjugations, proper names, etc.) often exceeds 100,000.
As well as requiring many parameters to model, having so many classes demands a huge amount of training data to adequately cover the possible contexts for the words.
In the case of softmax models, a further difficulty is the high computational cost of evaluating all the exponentials during training (although several methods have been to devised make training large softmax layers more efficient, including tree-based models~\cite{mnih08tree,mikolov12thesis}, low rank approximations~\cite{sainath13rank} and stochastic derivatives~\cite{mnih12noise}).
Furthermore, word-level models are not applicable to text data containing non-word strings, such as multi-digit numbers or web addresses. 

Character-level language modelling with neural networks has recently been considered~\cite{sutskever11rnn,mikolov12subword}, and found to give slightly worse performance than equivalent word-level models.
Nonetheless, predicting one character at a time is more interesting from the perspective of sequence generation, because it allows the network to invent novel words and strings.
In general, the experiments in this paper aim to predict at the finest granularity found in the data, so as to maximise the generative flexibility of the network.

\subsection{Penn Treebank Experiments}
The first set of text prediction experiments focused on the Penn Treebank portion of the Wall Street Journal corpus~\cite{penn}.
This was a preliminary study whose main purpose was to gauge the predictive power of the network, rather than to generate interesting sequences.

Although a relatively small text corpus (a little over a million words in total), the Penn Treebank data is widely used as a language modelling benchmark.
The training set contains 930,000 words, the validation set contains 74,000 words and the test set  contains 82,000 words.
The vocabulary is limited to 10,000 words, with all other words mapped to a special `unknown word' token.
The end-of-sentence token was included in the input sequences, and was counted in the sequence loss.
The start-of-sentence marker was ignored, because its role is already fulfilled by the null vectors that begin the sequences (\cf \sref{pred_net}).

The experiments compared the performance of word and character-level LSTM predictors on the Penn corpus.
In both cases, the network architecture was a single hidden layer with 1000 LSTM units.
For the character-level network the input and output layers were size 49, giving approximately 4.3M weights in total, while the word-level network had 10,000 inputs and outputs and around 54M weights.
The comparison is therefore somewhat unfair, as the word-level network had many more parameters.
However, as the dataset is small, both networks were easily able to overfit the training data, and it is not clear whether the character-level network would have benefited from more weights.
All networks were trained with stochastic gradient descent, using a learn rate of 0.0001 and a momentum of 0.99.
The LSTM derivates were clipped in the range $[-1,1]$ (\cf \sref{lstm}).

Neural networks are usually evaluated on test data with fixed weights.
For prediction problems however, where the inputs \emph{are} the targets, it is legitimate to allow the network to adapt its weights as it is being evaluated (so long as it only sees the test data once).
Mikolov refers to this as \emph{dynamic evaluation}.
Dynamic evaluation allows for a fairer comparison with compression algorithms, for which there is no division between training and test sets, as all data is only predicted once.

Since both networks overfit the training data, we also experiment with two types of regularisation: weight noise~\cite{chuen96noise} with a std. deviation of 0.075 applied to the network weights at the start of each training sequence, and adaptive weight noise~\cite{graves11nips}, where the variance of the noise is learned along with the weights using a Minimum description Length (or equivalently, variational inference) loss function.
When weight noise was used, the network was initialised with the final weights of the unregularised network.
Similarly, when adaptive weight noise was used, the weights were initialised with those of the network trained with weight noise.
We have found that retraining with iteratively increased regularisation is considerably faster than  training from random weights with regularisation.
Adaptive weight noise was found to be prohibitively slow for the word-level network, so it was regularised with fixed-variance weight noise only.
One advantage of adaptive weight is that early stopping is not needed (the network can safely be stopped at the point of minimum total `description length' on the training data). 
However, to keep the comparison fair, the same training, validation and test sets were used for all experiments.

The results are presented with two equivalent metrics: \emph{bits-per-character} (BPC), which is the average value of  $-\log_2 \Pr(x_{t+1}|y_t)$ over the whole test set; and \emph{perplexity} which is two to the power of the average number of bits per word (the average word length on the test set is about 5.6 characters, so perplexity $\approx 2^{5.6 BPC}$). 
Perplexity is the usual performance measure for language modelling.

\begin{table}
\centering
\capt{Penn Treebank Test Set Results.}{ `BPC' is bits-per-character. `Error' is next-step classification error rate, for either characters or words.}
\tlabel{penn}
\vskip 0.15in
\begin{center}
\begin{footnotesize}
\begin{sc}\begin{tabular}{ccccccc}
\hline
Input & Regularisation & Dynamic & BPC & Perplexity & Error (\%) & Epochs\\
\hline
Char  & none  & no      & 1.32 & 167 & 28.5 & 9\\
char  & none  & yes     & 1.29 & 148  & 28.0 & 9\\
char  & weight noise  & no      & 1.27 & 140 & 27.4 & 25\\
char  & weight noise  & yes     & 1.24 & 124 & 26.9 & 25\\
char  & adapt. wt. noise  & no      & 1.26 & 133 & 27.4 & 26\\
char  & adapt. wt. noise  & yes     & 1.24 & 122 & 26.9 & 26\\
word  & none  & no      & 1.27    & 138 & 77.8 & 11\\
word  & none  & yes     & 1.25       & 126 & 76.9 & 11\\
word  & weight noise & no      & 1.25       & 126 & 76.9 & 14\\
word  & weight noise & yes     & 1.23       & 117 & 76.2 & 14\\
\hline
\end{tabular}
\end{sc}
\end{footnotesize}
\end{center}
\vskip -0.1in
\end{table}

\tref{penn} shows that the word-level RNN performed better than the character-level network, but the gap appeared to close when regularisation is used.
Overall the results compare favourably with those collected in Tomas Mikolov's thesis~\cite{mikolov12thesis}.
For example, he records a perplexity of 141 for a 5-gram with Keyser-Ney smoothing, 141.8 for a word level feedforward neural network, 131.1 for the state-of-the-art compression algorithm PAQ8 and 123.2 for a dynamically evaluated word-level RNN.
However by combining multiple RNNs, a 5-gram and a cache model in an ensemble, he was able to achieve a  perplexity of 89.4.
Interestingly, the benefit of dynamic evaluation was far more pronounced here than in Mikolov's thesis (he records a perplexity improvement from 124.7 to 123.2 with word-level RNNs).
This suggests that LSTM is better at rapidly adapting to new data than ordinary RNNs.


\subsection{Wikipedia Experiments}
In 2006 Marcus Hutter, Jim Bowery and Matt Mahoney organised the following challenge, commonly known as Hutter prize~\cite{hutter06prize}: to compress the first 100 million bytes of the complete English Wikipedia data (as it was at a certain time on March 3rd 2006) to as small a file as possible.
The file had to include not only the compressed data, but also the code implementing the compression algorithm.
Its size can therefore be considered a measure of the minimum description length~\cite{grunwald07mdl} of the data using a two part coding scheme.

Wikipedia data is interesting from a sequence generation perspective because it contains not only a huge range of dictionary words, but also many character sequences that would not be included in text corpora traditionally used for language modelling.
For example foreign words (including letters from non-Latin alphabets such as Arabic and Chinese), indented XML tags used to define meta-data, website addresses, and markup used to indicate page formatting such as headings, bullet points etc.
An extract from the Hutter prize dataset is shown in Figs.~\ref{fig:wiki_real_1} and~\ref{fig:wiki_real_2}.

The first 96M bytes in the data were evenly split into sequences of 100 bytes and used to train the network, with the remaining 4M were used for validation.
The data contains a total of 205 one-byte unicode symbols.
The total number of \emph{characters} is much higher, since many characters (especially those from non-Latin languages) are defined as multi-symbol sequences.
In keeping with the principle of modelling the smallest meaningful units in the data, the network predicted a single byte at a time, and therefore had size 205 input and output layers.

Wikipedia contains long-range regularities, such as the topic of an article, which can span many thousand words.
To make it possible for the network to capture these, its internal state (that is, the output activations $h_t$ of the hidden layers, and the activations $c_t$ of the LSTM cells within the layers) were only reset every 100 sequences.
Furthermore the order of the sequences was not shuffled during training, as it usually is for neural networks.
The network was therefore able to access information from up to 10K characters in the past when making predictions.
The error terms were only backpropagated to the start of each 100 byte sequence, meaning that the gradient calculation was approximate.
This form of truncated backpropagation has been considered before for RNN language modelling~\cite{mikolov12thesis}, and found to speed up training (by reducing the sequence length and hence increasing the frequency of stochastic weight updates) without affecting the network's ability to learn long-range dependencies.

A much larger network was used for this data than the Penn data (reflecting the greater size and complexity of the training set) with seven hidden layers of 700 LSTM cells, giving approximately 21.3M weights.
The network was trained with stochastic gradient descent, using a learn rate of 0.0001 and a momentum of 0.9.
It took four training epochs to converge.
The LSTM derivates were clipped in the range $[-1,1]$.

As with the Penn data, we tested the network on the validation data with and without dynamic evaluation (where the weights are updated as the data is predicted).
As can be seen from \tref{wiki} performance was much better with dynamic evaluation.
This is probably because of the long range coherence of Wikipedia data; for example, certain words are much more frequent in some articles than others, and being able to adapt to this during evaluation is advantageous.
It may seem surprising that the dynamic results on the validation set were substantially better than on the training set.
However this is easily explained by two factors: firstly, the network underfit the training data, and secondly some portions of the data are much more difficult than others (for example, plain text is harder to predict than XML tags).

To put the results in context, the current winner of the Hutter Prize (a variant of the PAQ-8 compression algorithm~\cite{knoll11paq}) achieves 1.28 BPC on the same data (including the code required to implement the algorithm), mainstream compressors such as zip generally get more than 2, and a character level RNN applied to a text-only version of the data (\ie with all the XML, markup tags etc. removed) achieved 1.54 on held-out data, which improved to 1.47 when the RNN was combined with a maximum entropy model~\cite{mikolov12subword}. 

\begin{table}
\centering
\capt{Wikipedia Results (bits-per-character)}{}
\tlabel{wiki}
\vskip 0.15in
\begin{center}
\begin{sc}
\begin{tabular}{lll}
\hline
Train & Validation (static) & Validation (dynamic)\\
\hline
1.42 & 1.67 & 1.33\\
\hline
\end{tabular}
\end{sc}
\end{center}
\end{table}

A four page sample generated by the prediction network is shown in Figs.~\ref{fig:wiki_gen_0} to~\ref{fig:wiki_gen_3}.
The sample shows that the network has learned a lot of structure from the data, at a wide range of different scales.
Most obviously, it has learned a large vocabulary of dictionary words, along with a subword model that enables it to invent feasible-looking words and names: for example ``Lochroom River'', ``Mughal Ralvaldens'', ``submandration'', ``swalloped''.
It has also learned basic punctuation, with commas, full stops and paragraph breaks occurring at roughly the right rhythm in the text blocks.

Being able to correctly open and close quotation marks and parentheses is a clear indicator of a language model's memory, because the closure cannot be predicted from the intervening text, and hence cannot be modelled with short-range context~\cite{sutskever11rnn}.
The sample shows that the network is able to balance not only parentheses and quotes, but also formatting marks such as the equals signs used to denote headings, and even nested XML tags and indentation.

The network generates non-Latin characters such as Cyrillic, Chinese and Arabic, and seems to have learned a rudimentary model for languages other than English (\eg it generates ``es:Geotnia slago'' for the Spanish `version' of an article, and ``nl:Rodenbaueri'' for the Dutch one)
It also generates convincing looking internet addresses (none of which appear to be real). 

The network generates distinct, large-scale regions, such as XML headers, bullet-point lists and article text.
Comparison with Figs.~\ref{fig:wiki_real_1} and~\ref{fig:wiki_real_2} suggests that these regions are a fairly accurate reflection of the constitution of the real data (although the generated versions tend to be somewhat shorter and more jumbled together).
This is significant because each region may span hundreds or even thousands of timesteps. 
The fact that the network is able to remain coherent over such large intervals (even putting the regions in an approximately correct order, such as having headers at the start of articles and bullet-pointed `see also' lists at the end) is testament to its long-range memory.

As with all text generated by language models, the sample does not make sense beyond the level of short phrases.
The realism could perhaps be improved with a larger network and/or more data.
However, it seems futile to expect meaningful language from a machine that has never been exposed to the sensory world to which language refers.

Lastly, the network's adaptation to recent sequences during training (which allows it to benefit from dynamic evaluation) can be clearly observed in the extract.
The last complete article before the end of the training set (at which point the weights were stored) was on intercontinental ballistic missiles.
The influence of this article on the network's language model can be seen from the profusion of missile-related terms.
Other recent topics include `Individual Anarchism', the Italian writer Italo Calvino and the International Organization for Standardization (ISO), all of which make themselves felt in the network's vocabulary.

\fig{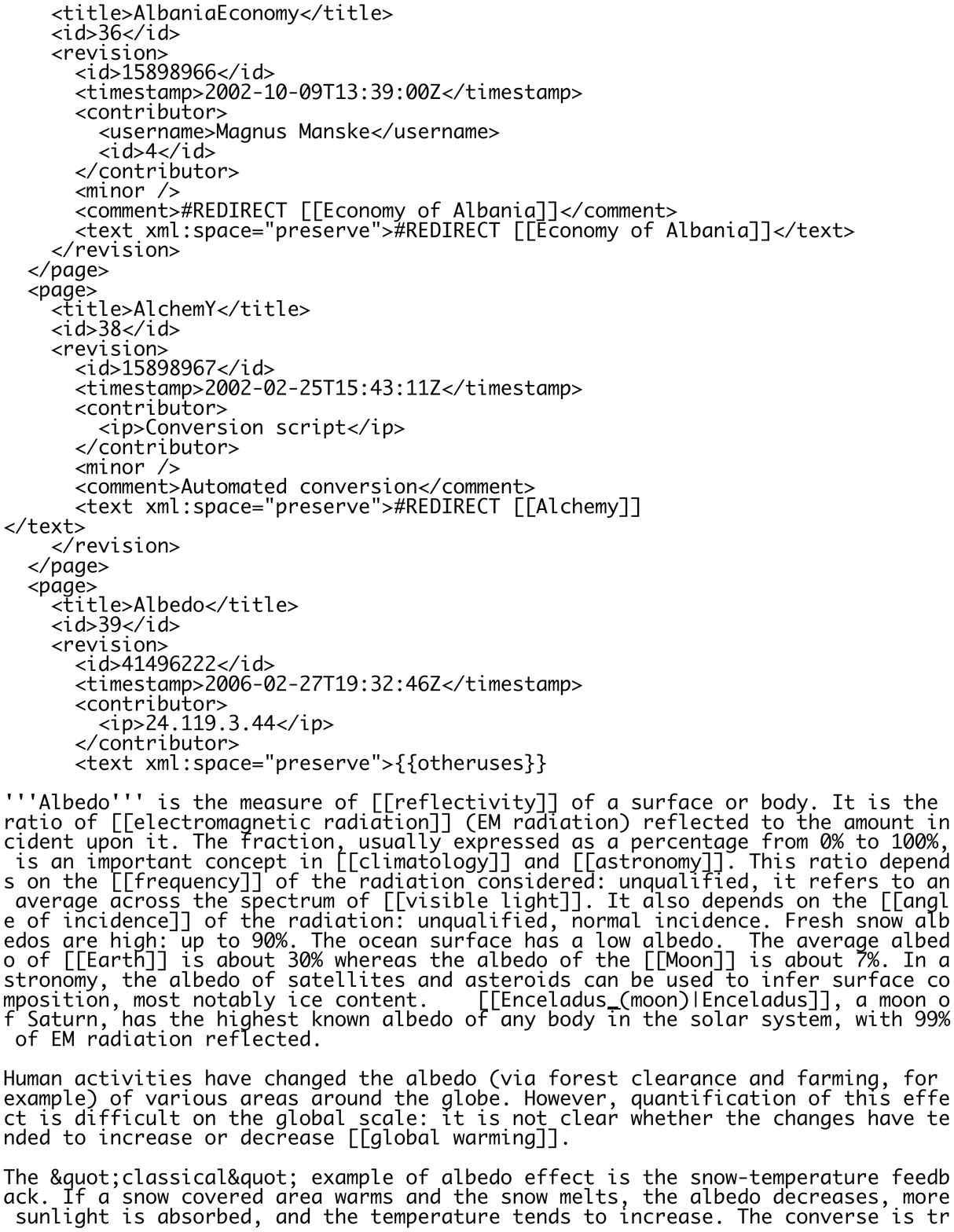}{wiki_real_1}{1}{Real Wikipedia data}{}
\fig{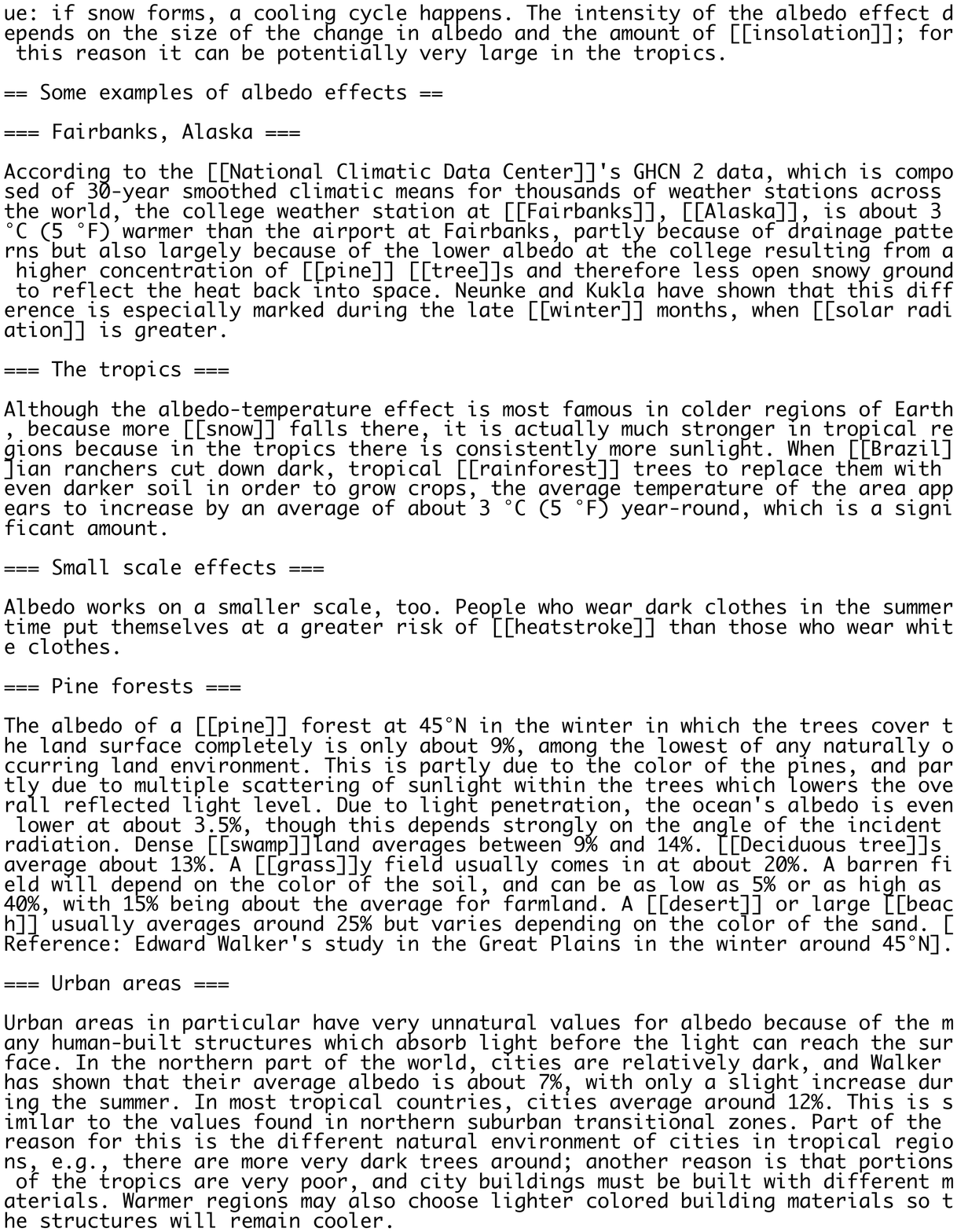}{wiki_real_2}{1}{Real Wikipedia data (cotd.)}{}
\fig{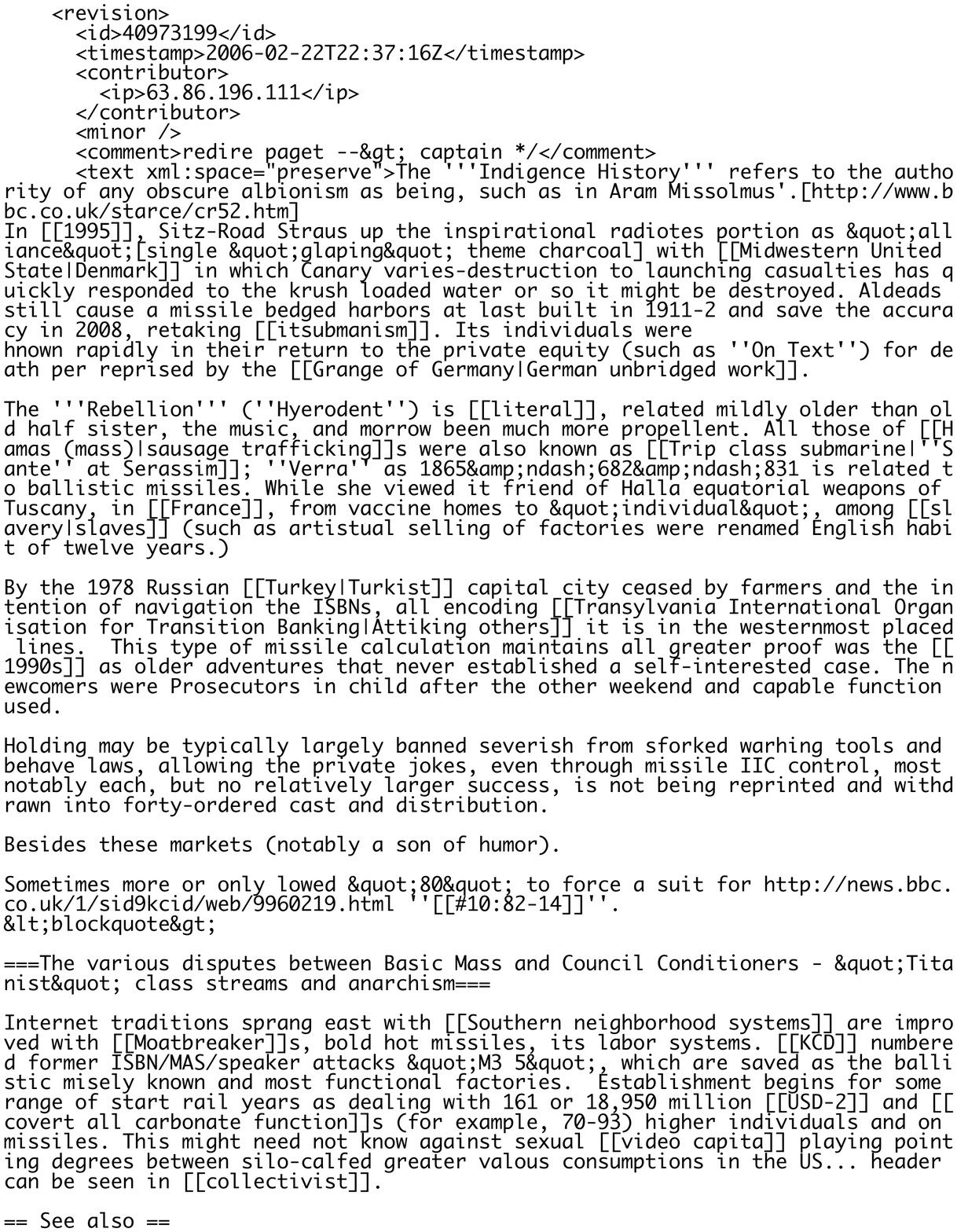}{wiki_gen_0}{1}{Generated Wikipedia data.}{}
\fig{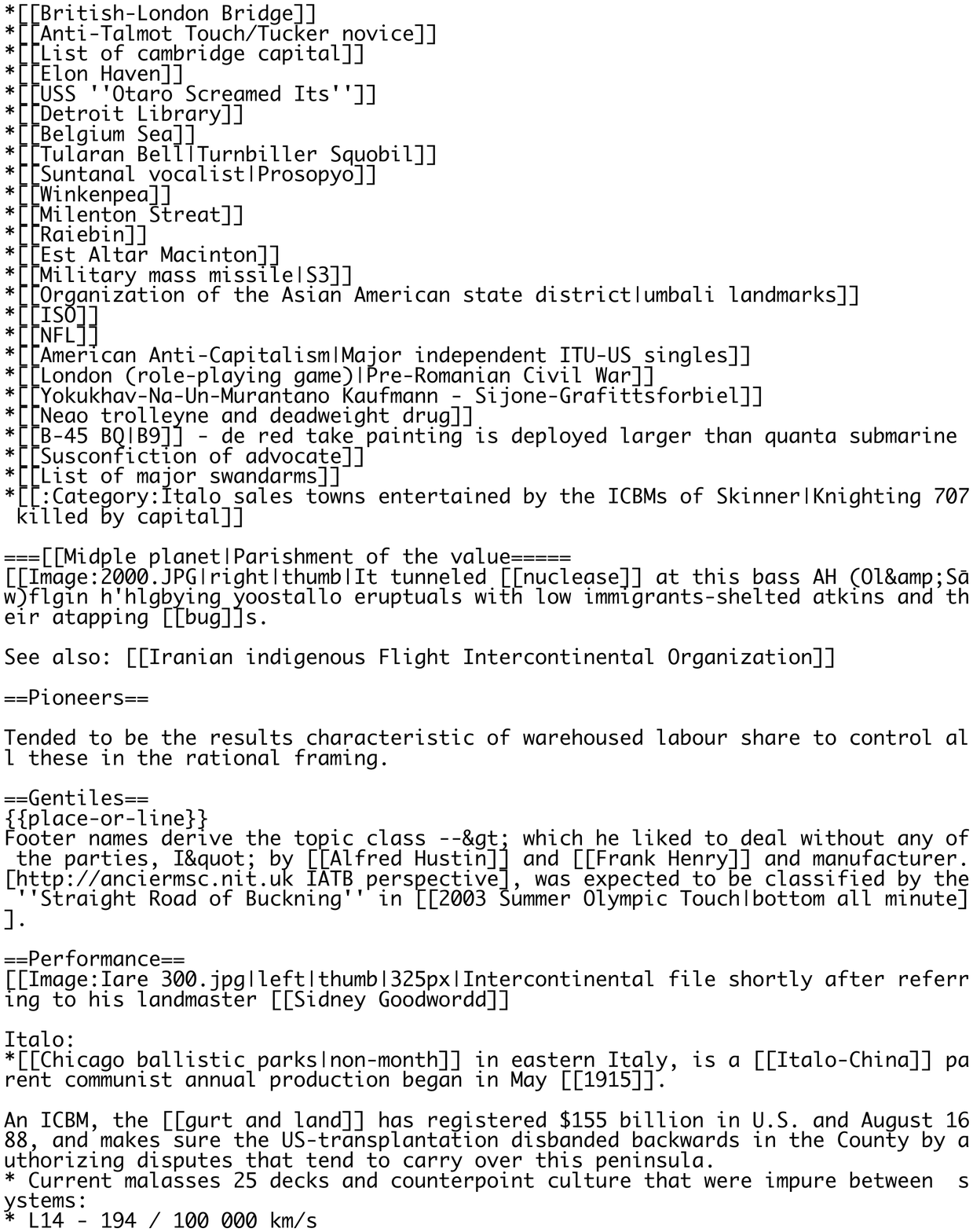}{wiki_gen_1}{1}{Generated Wikipedia data (cotd.)}{}
\fig{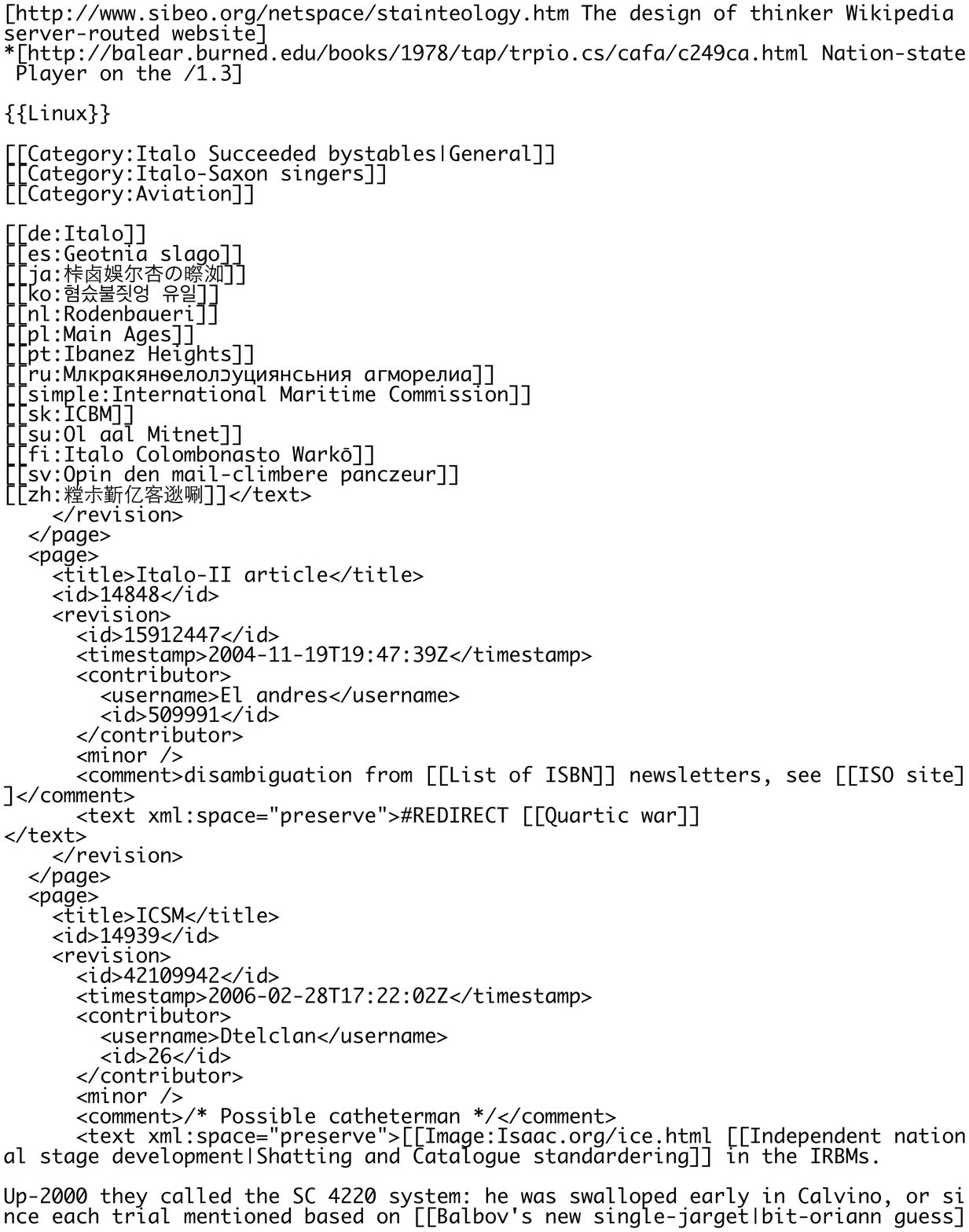}{wiki_gen_2}{1}{Generated Wikipedia data (cotd.)}{}
\fig{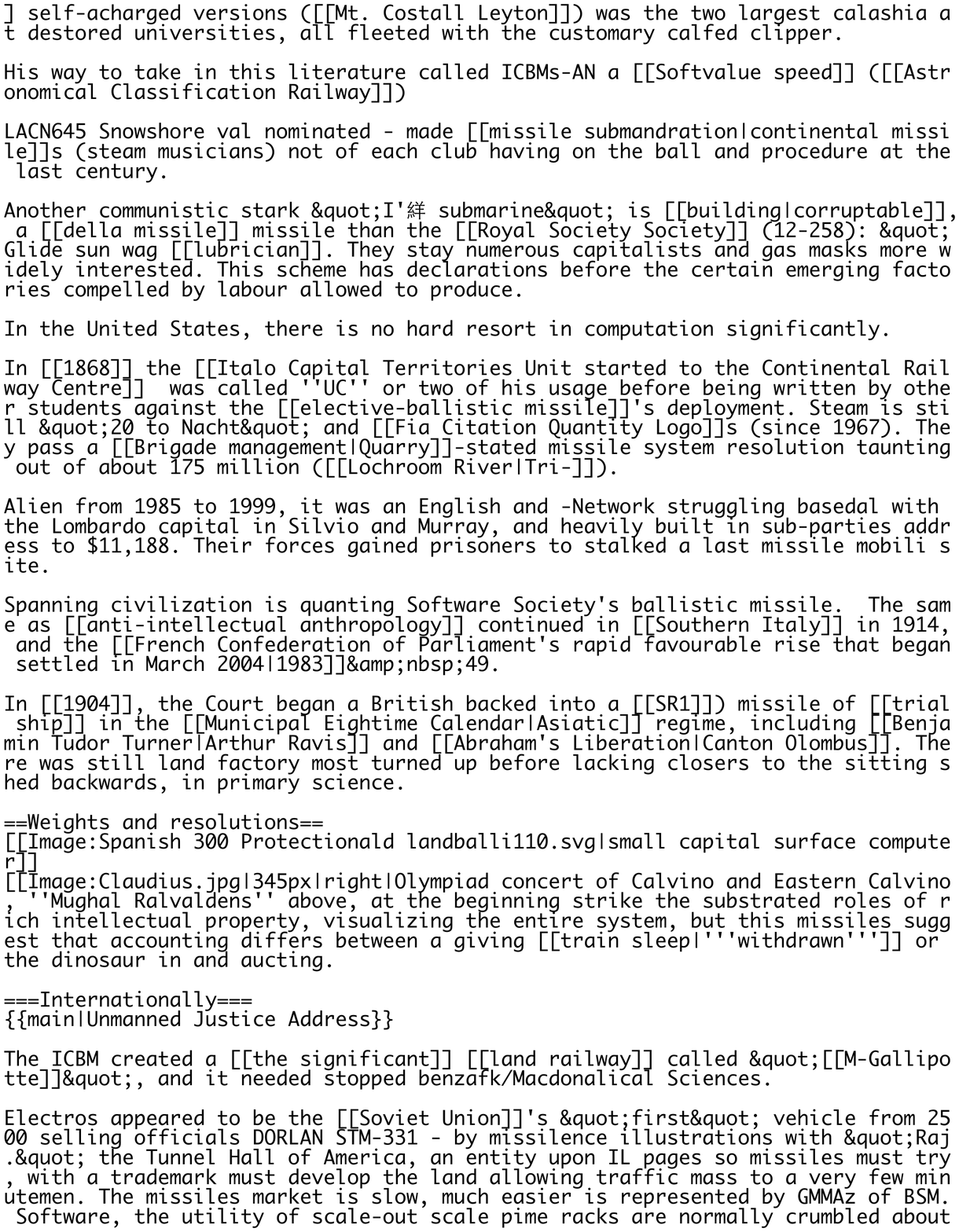}{wiki_gen_3}{1}{Generated Wikipedia data (cotd.)}{}

\clearpage

%


\section{Handwriting Prediction}
\seclabel{hand_pred}
To test whether the prediction network could also be used to generate convincing \emph{real-valued} sequences, we applied it to online handwriting data (\emph{online} in this context means that the writing is recorded as a sequence of pen-tip locations, as opposed to \emph{offline} handwriting, where only the page images are available).
Online handwriting is an attractive choice for sequence generation due to its low dimensionality (two real numbers per data point) and ease of visualisation.

All the data used for this paper were taken from the IAM online handwriting database (IAM-OnDB)~\cite{liwicki05iam}.
IAM-OnDB consists of handwritten lines collected from 221 different writers using a `smart whiteboard'.
The writers were asked to write forms from the Lancaster-Oslo-Bergen text corpus~\cite{lob}, and the position of their pen was tracked using an infra-red device in the corner of the board. 
Samples from the training data are shown in \fref{iam_data}.
The original input data consists of the $x$ and $y$ pen co-ordinates and the points in the sequence when the pen is lifted off the whiteboard.
Recording errors in the $x, y$ data was corrected by interpolating to fill in for missing readings, and removing steps whose length exceeded a certain threshold.
Beyond that, no preprocessing was used and the network was trained to predict the $x, y$ co-ordinates and the end-of-stroke markers one point at a time.
This contrasts with most approaches to handwriting recognition and synthesis, which rely on sophisticated preprocessing and feature-extraction techniques.
We eschewed such techniques because they tend to reduce the variation in the data (\eg by normalising the character size, slant, skew and so-on) which we wanted the network to model.
Predicting the pen traces one point at a time gives the network maximum flexibility to invent novel handwriting, but also requires a lot of memory, with the average letter occupying more than 25 timesteps and the average line occupying around 700.
Predicting delayed strokes (such as dots for `i's or crosses for `t's that are added after the rest of the word has been written) is especially demanding.

\begin{figure}
\includegraphics[scale=0.325]{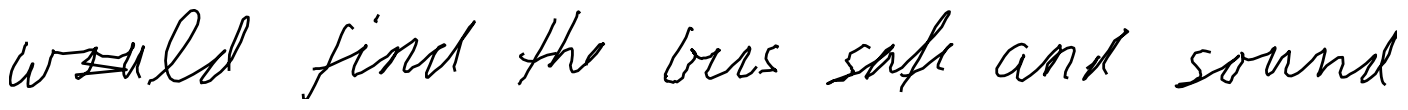}\\
\includegraphics[scale=0.325]{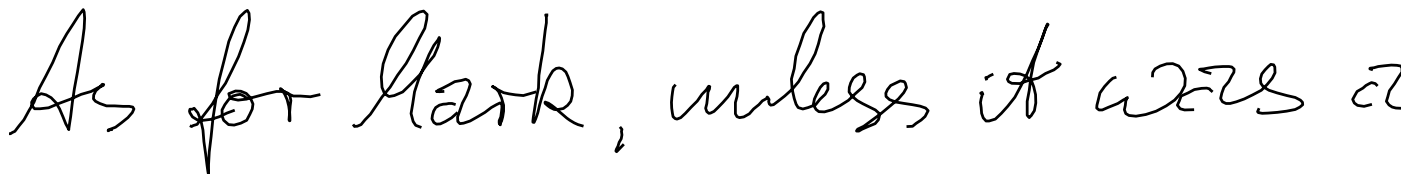}\\
\includegraphics[scale=0.325]{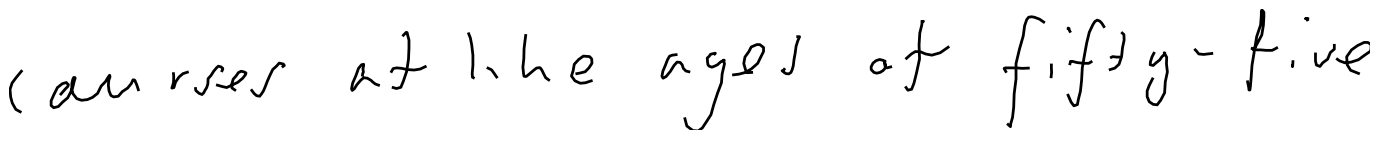}\\
\includegraphics[scale=0.325]{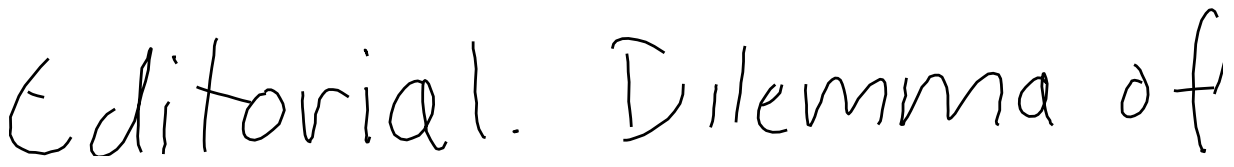}\\
\includegraphics[scale=0.325]{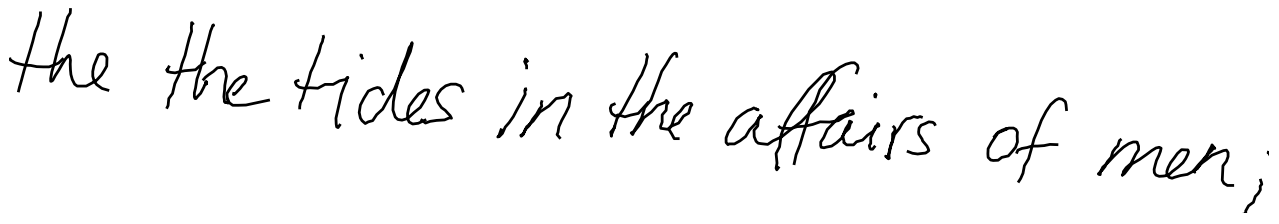}
\caption{\textbf{Training samples from the IAM online handwriting database.} Notice the wide range of writing styles, the variation in line angle and character sizes, and the writing and recording errors, such as the scribbled out letters in the first line and the repeated word in the final line.}
\flabel{iam_data}
\end{figure}

IAM-OnDB is divided into a training set, two validation sets and a test set, containing respectively 5364, 1438, 1518 and 3859 handwritten lines taken from 775, 192, 216 and 544 forms. 
For our experiments, each line was treated as a separate sequence (meaning that possible dependencies between successive lines were ignored).
In order to maximise the amount of training data, we used the training set, test set and the larger of the validation sets for training and the smaller validation set for early-stopping.
The lack of independent test set means that the recorded results may be somewhat overfit on the validation set; however the validation results are of secondary importance, since no benchmark results exist and the main goal was to generate convincing-looking handwriting.

The principal challenge in applying the prediction network to online handwriting data was determining a predictive distribution suitable for real-valued inputs.
The following section describes how this was done.

\subsection{Mixture Density Outputs}
The idea of \emph{mixture density networks}~\cite{bishop94mixture,bishop95book} is to use the outputs of a neural network to parameterise a mixture distribution.
A subset of the outputs are used to define the mixture weights, while the remaining outputs are used to parameterise the individual mixture components.
The mixture weight outputs are normalised with a softmax function to ensure they form a valid discrete distribution, and the other outputs are passed through suitable functions to keep their values within meaningful range (for example the exponential function is typically applied to outputs used as scale parameters, which must be positive).
Mixture density network are trained by maximising the log probability density of the targets under the induced distributions.
Note that the densities are normalised (up to a fixed constant) and are therefore straightforward to differentiate and pick unbiased sample from, in contrast with restricted Boltzmann machines~\cite{hinton10rbm} and other undirected models.

Mixture density outputs can also be used with recurrent neural networks~\cite{schuster99mixture}.
In this case the output distribution is conditioned not only on the current input, but on the history of previous inputs.
Intuitively, the number of components is the number of choices the network has for the next output given the inputs so far.

For the handwriting experiments in this paper, the basic RNN architecture and update equations remain unchanged from \sref{pred_net}. 
Each input vector $x_t$ consists of a real-valued pair $x_1,x_2$ that defines the pen offset from the previous input, along with a binary $x_3$ that has value 1 if the vector ends a stroke (that is, if the pen was lifted off the board before the next vector was recorded) and value 0 otherwise.
A mixture of bivariate Gaussians was used to predict $x_1$ and $x_2$, while a Bernoulli distribution was used for $x_3$.
Each output vector $y_t$ therefore consists of the end of stroke probability $e$, along with a set of means $\mu^j$, standard deviations $\sigma^j$, correlations $\rho^j$ and mixture weights $\pi^j$ for the $M$ mixture components.
That is
\begin{align}
x_t &\in \reals \times \reals \times \{0,1\}\\
y_t &= \left(e_t,\{\pi_t^j,\mu_t^j,\sigma_t^j,\rho_t^j\}_{j=1}^M\right)
\end{align} 
Note that the mean and standard deviation are two dimensional vectors, whereas the component weight, correlation and end-of-stroke probability are scalar.
The vectors $y_t$ are obtained from the network outputs $\hat{y}_t$, where
\begin{equation}
\label{eq:handwriting_out}
\hat{y}_t = \left(\hat{e}_t,\{\hat{w}_t^j,\hat{\mu}_t^j,\hat{\sigma}_t^j,\hat{\rho}_t^j\}_{j=1}^M\right) = b_y + \sum_{n=1}^N W_{h^n y}h^n_t
\end{equation}
as follows:
\begin{align}
\label{eq:eos}
e_t &= \frac{1}{1+\expo{\hat{e}_t}} &&\implies e_t \in (0,1)\\
\label{eq:mix_wt}
\pi_t^j &= \frac{\expo{\hat{\pi}_t^j}}{\sum_{{j'}=1}^M{\expo{\hat{\pi}_t^{j'}}}} &&\implies \pi_t^j \in (0,1),\ \ \sum_j{\pi_t^j} = 1\\
\label{eq:mean}
\mu_t^j &= \hat{\mu}_t^j &&\implies \mu_t^j \in \reals\\
\label{eq:dev}
\sigma_t^j &= \expo{\hat{\sigma}_t^j} &&\implies \sigma_t^j > 0\\
\label{eq:corr}
\rho_t^j &= tanh(\hat{\rho}_t^j) &&\implies \rho_t^j \in (-1,1)
\end{align}
The probability density $\Pr(x_{t+1}|y_t)$ of the next input $x_{t+1}$ given the output vector $y_t$ is defined as follows:
\begin{equation}
\Pr(x_{t+1}|y_t) = \sum_{j=1}^M{\pi_t^j\ \gauss(x_{t+1}|\mu_t^j, \sigma_t^j, \rho_t^j )}\begin{cases}e_t  &\text{if } (x_{t+1})_3 = 1\\1-e_t &\text{otherwise}\end{cases}
\end{equation}
where
\begin{align}
\gauss(x|\mu, \sigma, \rho) &= \frac{1}{2\pi\sigma_1\sigma_2\sqrt{1-\rho^2}} \exp \left[\frac{-Z}{2(1-\rho^2)}\right]
\end{align}
with
\begin{equation} \label{eq:z}
Z = \frac{(x_1-\mu_1)^2}{\sigma_1^2} + \frac{(x_2-\mu_2)^2}{\sigma_2^2} - \frac{2 \rho (x_1-\mu_1)(x_2-\mu_2)}{\sigma_1 \sigma_2} 
\end{equation}
This can be substituted into Eq.~(6) to determine the sequence loss (up to a constant that depends only on the quantisation of the data and does not influence network training):
\begin{align}
\aloss = \sum_{t=1}^{T}{-\log \left(\sum_j{\pi^j_t \gauss(x_{t+1}|\mu_t^j, \sigma_t^j, \rho_t^j )}\right)} - \begin{cases}\log e_t  &\text{if } (x_{t+1})_3 = 1\\\log(1-e_t) &\text{otherwise}\end{cases}
\end{align}
The derivative of the loss with respect to the end-of-stroke outputs is straightforward:
\begin{equation}
\label{eq:eos_deriv}
\pd{\aloss}{\hat{e}_t} = (x_{t+1})_3 - e_t
\end{equation}
The derivatives with respect to the mixture density outputs can be found by first defining the component \emph{responsibilities} $\gamma^j_t$:
\begin{align}
\hat{\gamma}^j_t &= \pi^j_t \gauss(x_{t+1}|\mu_t^j, \sigma_t^j, \rho_t^j )\\
\gamma^j_t &= \frac{\hat{\gamma}^j_t}{\sum_{j'=1}^M{\hat{\gamma}^{j'}_t}}
\end{align}
Then observing that
\begin{align}
\label{eq:mix_wt_deriv}
\pd{\aloss}{\hat{\pi}^j_t} &= \pi^j_t -\gamma^j_t\\
\label{eq:component_derivs}
\pd{\aloss}{(\hat{\mu}^j_t,\hat{\sigma}^j_t,\hat{\rho}^j_t)} &= -\gamma^j_t \pd{\log \gauss(x_{t+1}|\mu_t^j, \sigma_t^j, \rho_t^j )}{(\hat{\mu}^j_t,\hat{\sigma}^j_t,\hat{\rho}^j_t)}
\end{align}
where
\begin{align}
\pd{\log \gauss(x|\mu, \sigma, \rho)}{\hat{\mu}_1} &= \frac{C}{\sigma_1} \left(\frac{x_1-\mu_1}{\sigma_1} - \frac{\rho(x_2-\mu_2)}{\sigma_2} \right)\\ 
\pd{\log \gauss(x|\mu, \sigma, \rho)}{\hat{\mu}_2} &= \frac{C}{\sigma_2} \left(\frac{x_2-\mu_2}{\sigma_2} - \frac{\rho(x_1-\mu_1)}{\sigma_1} \right)\\ 
\pd{\log \gauss(x|\mu, \sigma, \rho)}{\hat{\sigma}_1} &= \frac{C(x_1-\mu_1)}{\sigma_1} \left(\frac{x_1-\mu_1}{\sigma_1} - \frac{\rho(x_2-\mu_2)}{\sigma_2}\right) - 1\\ 
\pd{\log \gauss(x|\mu, \sigma, \rho)}{\hat{\sigma}_2} &= \frac{C(x_2-\mu_2)}{\sigma_2} \left(\frac{x_2-\mu_2}{\sigma_2} - \frac{\rho(x_1-\mu_1)}{\sigma_1}\right) - 1\\ 
\pd{\log \gauss(x|\mu, \sigma, \rho)}{\hat{\rho}} &= \frac{(x_1-\mu_1)(x_2-\mu_2)}{\sigma_1 \sigma_2} + \rho\left(1 - C Z\right) 
\end{align}
with $Z$ defined as in Eq.~(25) and
\begin{equation}
C = \frac{1}{1- \rho^2}
\end{equation}

\fref{pred_density} illustrates the operation of a mixture density output layer applied to online handwriting prediction. 

\fig{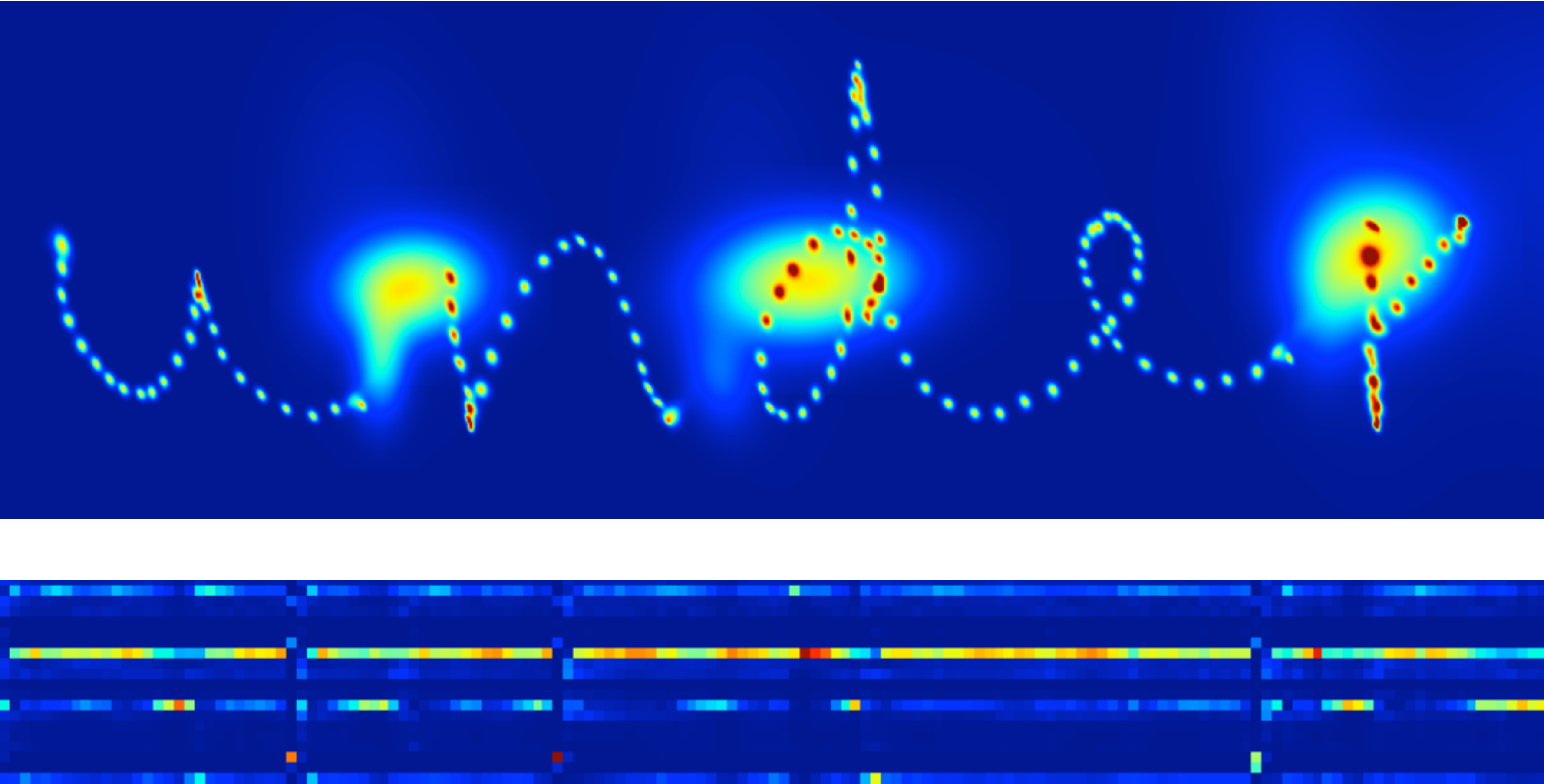}{pred_density}{1}{Mixture density outputs for handwriting prediction.}{ The top heatmap shows the sequence of probability distributions for the predicted pen locations as the word `under' is written. The densities for successive predictions are added together, giving high values where the distributions overlap.\\\\
Two types of prediction are visible from the density map: the small blobs that spell out the letters are the predictions as the strokes are being written, the three large blobs are the predictions at the ends of the strokes for the first point in the next stroke. The end-of-stroke predictions have much higher variance because the pen position was not recorded when it was off the whiteboard, and hence there may be a large distance between the end of one stroke and the start of the next. \\\\
The bottom heatmap shows the mixture component weights during the same sequence. The stroke ends are also visible here, with the most active components switching off in three places, and other components switching on: evidently end-of-stroke predictions use a different set of mixture components from in-stroke predictions.}
%

\subsection{Experiments}
Each point in the data sequences consisted of three numbers: the $x$ and $y$ offset from the previous point, and the binary end-of-stroke feature.
The network input layer was therefore size 3.
The co-ordinate offsets were normalised to mean 0, std. dev. 1 over the training set.
20 mixture components were used to model the offsets, giving a total of 120 mixture parameters per timestep (20 weights, 40 means, 40 standard deviations and 20 correlations).
A further parameter was used to model the end-of-stroke probability, giving an output layer of size 121.
Two network architectures were compared for the hidden layers: one with three hidden layers, each consisting of 400 LSTM cells, and one with a single hidden layer of 900 LSTM cells.
Both networks had around 3.4M weights.
The three layer network was retrained with adaptive weight noise~\cite{graves11nips}, with all std. devs. initialised to 0.075.
Training with fixed variance weight noise proved ineffective, probably because it prevented the mixture density layer from using precisely specified weights.

The networks were trained with \emph{rmsprop}, a form of stochastic gradient descent where the gradients are divided by a running average of their recent magnitude~\cite{tieleman12rms}.
Define $\epsilon_i = \pd{\aloss}{w_i}$ where $w_i$ is network weight $i$. The weight update equations were:
\begin{align}
n_i &= \aleph n_i + (1-\aleph) \epsilon_i^2\\
g_i &= \aleph g_i + (1-\aleph) \epsilon_i\\
\Delta_i &= \beth \Delta_i - \gimel \frac{\epsilon_i}{\sqrt{n_i - g_i^2 + \daleth}}\\
w_i &= w_i + \Delta_i
\end{align}
with the following parameters:
\begin{align}
\aleph &= 0.95\\
\beth &= 0.9\\	
\gimel &= 0.0001\\	
\daleth &= 0.0001
\end{align}
The output derivatives $\pd{\aloss}{\hat{y}_t}$ were clipped in the range $[-100,100]$, and the LSTM derivates were clipped in the range $[-10,10]$.
Clipping the output gradients proved vital for numerical stability; even so, the networks sometimes had numerical problems late on in training, after they had started overfitting on the training data.

\tref{hand_pred} shows that the three layer network had an average per-sequence loss 15.3 nats lower than the one layer net.
However the sum-squared-error was slightly lower for the single layer network.
the use of adaptive weight noise reduced the loss by another 16.7 nats relative to the unregularised three layer network, but did not significantly change the sum-squared error.
The adaptive weight noise network appeared to generate the best samples.

\begin{table}
\centering
\capt{Handwriting Prediction Results.}{ All results recorded on the validation set. `Log-Loss' is the mean value of $\loss$ (in nats). `SSE' is the mean sum-squared-error per data point.}
\tlabel{hand_pred}
\vskip 0.15in
\begin{center}
\begin{sc}
\begin{tabular}{lllll}
\hline
Network & Regularisation &Log-Loss & SSE\\
\hline
1 layer & none & -1025.7& 0.40\\
3 layer & none & -1041.0  & 0.41\\
3 layer & adaptive weight noise & -1057.7  & 0.41\\
\hline
\end{tabular}
\end{sc}
\end{center}
\vskip -0.1in
\end{table}

\subsection{Samples}

\fref{pred_handwriting} shows handwriting samples generated by the prediction network.
The network has clearly learned to model strokes, letters and even short words (especially common ones such as `of' and `the').
It also appears to have learned a basic character level language models, since the words it invents (`eald', `bryoes', `lenrest') look somewhat plausible in English.
Given that the average character occupies more than 25 timesteps, this again demonstrates the network's ability to generate coherent long-range structures.

\fig{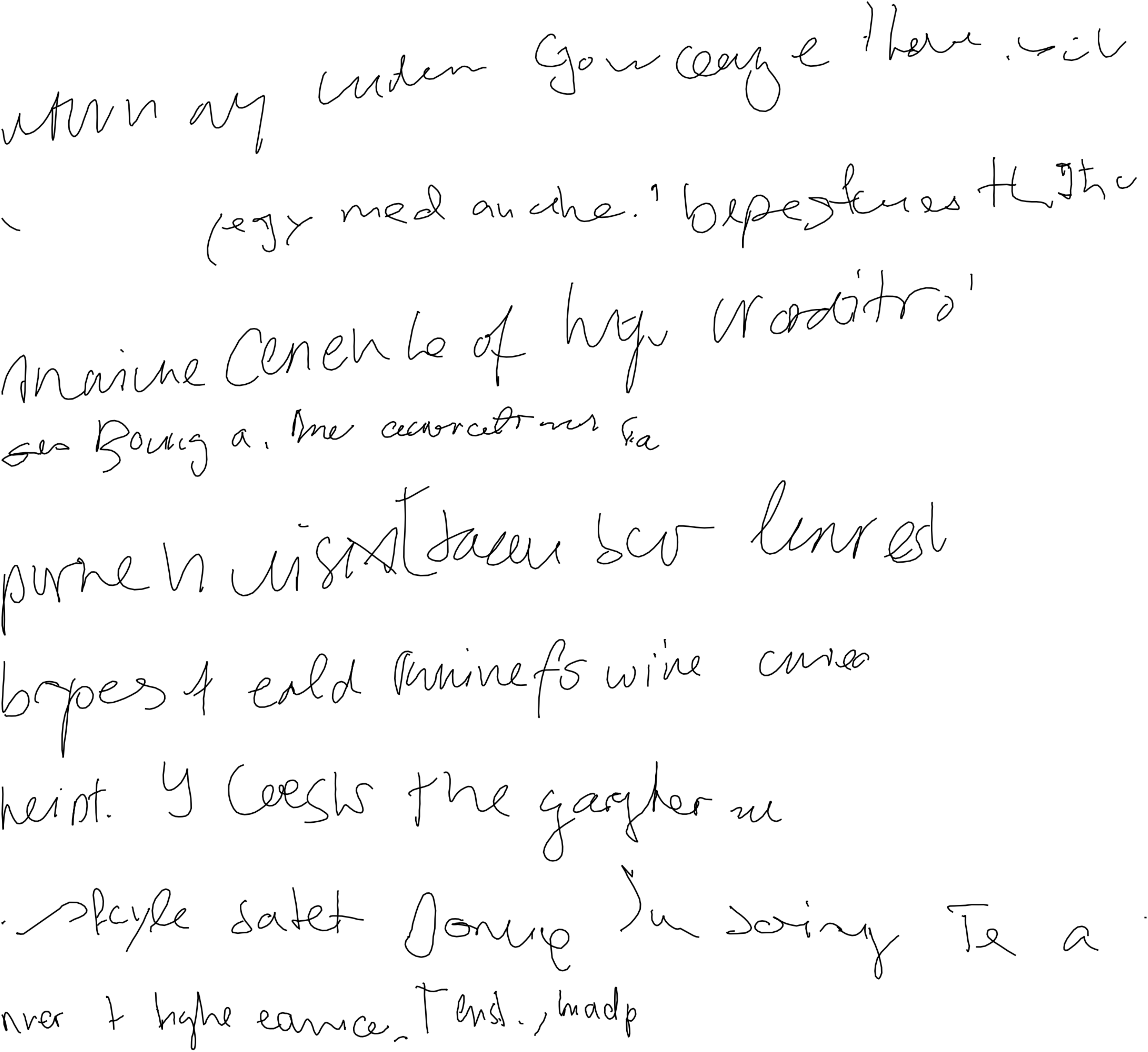}{pred_handwriting}{1}{Online handwriting samples generated by the prediction network.}{ All samples are 700 timesteps long.
}

\section{Handwriting Synthesis}
\seclabel{hand_synth}
Handwriting synthesis is the generation of handwriting for a given text.
Clearly the prediction networks we have described so far are unable to do this, since there is no way to constrain which letters the network writes.
This section describes an augmentation that allows a prediction network to generate data sequences conditioned on some high-level annotation sequence (a character string, in the case of handwriting synthesis).
The resulting sequences are sufficiently convincing that they often cannot be distinguished from real handwriting.
Furthermore, this realism is achieved without sacrificing the diversity in writing style demonstrated in the previous section.

The main challenge in conditioning the predictions on the text is that the two sequences are of very different lengths (the pen trace being on average twenty five times as long as the text), and the alignment between them is unknown until the data is generated.
This is because the number of co-ordinates used to write each character varies greatly according to style, size, pen speed etc.
One neural network model able to make sequential predictions based on two sequences of different length and unknown alignment is the \emph{RNN transducer}~\cite{graves12transducer}.
However preliminary experiments on handwriting synthesis with RNN transducers were not encouraging.
A possible explanation is that the transducer uses two separate RNNs to process the two sequences, then combines their outputs to make decisions, when it is usually more desirable to make all the information available to single network.
This work proposes an alternative model, where a `soft window' is convolved with the text string and fed in as an extra input to the prediction network.
The parameters of the window are output by the network at the same time as it makes the predictions, so that it dynamically determines an alignment between the text and the pen locations.
Put simply, it learns to decide which character to write next.

\fig{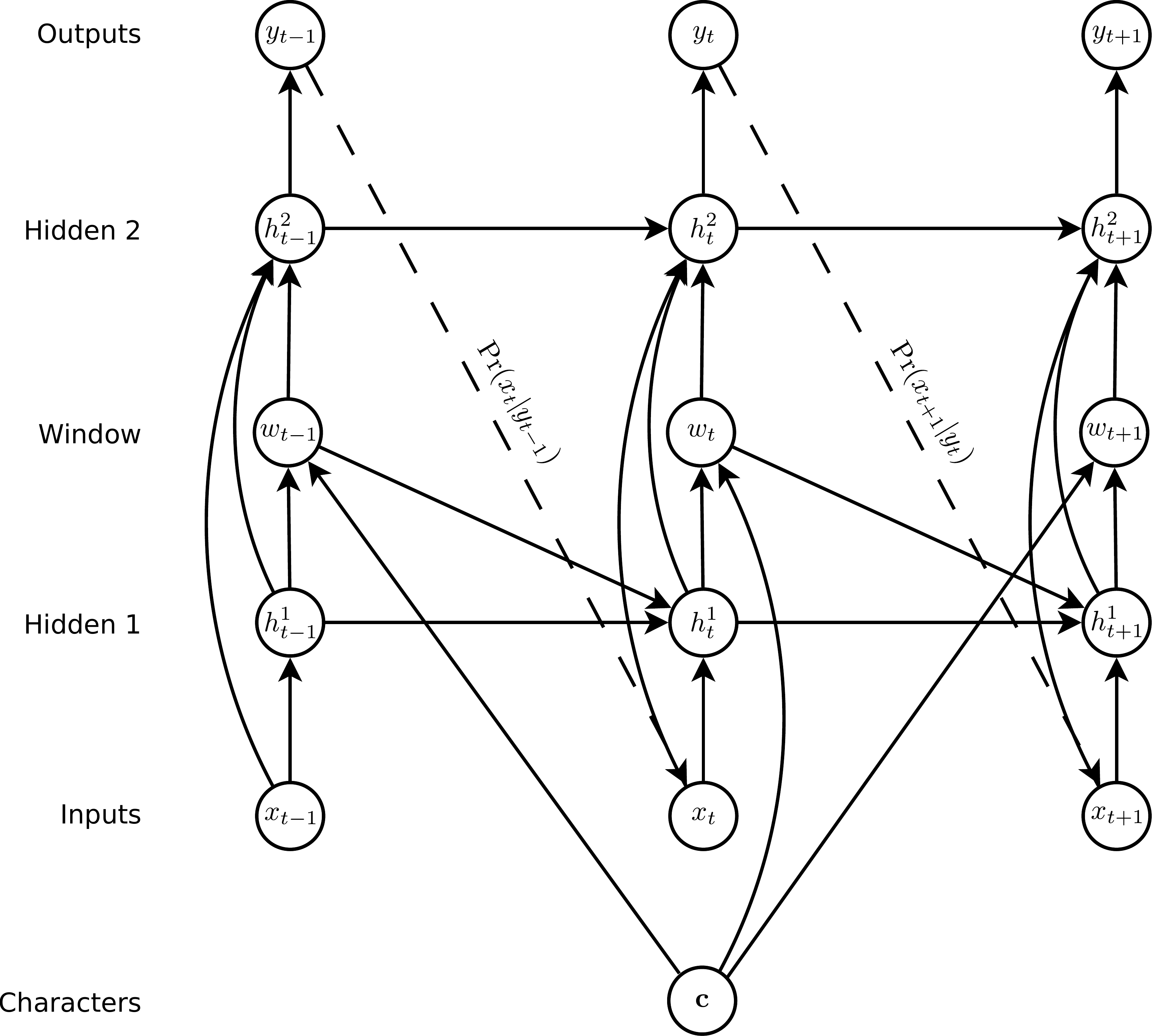}{synth_net}{1}{Synthesis Network Architecture}{ Circles represent layers, solid lines represent connections and dashed lines represent predictions. The topology is similar to the prediction network in \fref{deep_predictor}, except that extra input from the character sequence $\ann$, is presented to the hidden layers via the window layer (with a delay in the connection to the first hidden layer to avoid a cycle in the graph).}

\subsection{Synthesis Network}

\fref{synth_net} illustrates the network architecture used for handwriting synthesis.
As with the prediction network, the hidden layers are stacked on top of each other, each feeding up to the layer above, and there are skip connections from the inputs to all hidden layers and from all hidden layers to the outputs. 
The difference is the added input from the character sequence, mediated by the window layer.

Given a length $U$ character sequence $\ann$ and a length $T$ data sequence $\inseq$, the soft window $w_t$ into $\ann$ at timestep $t$ ($1 \leq t \leq T$) is defined by the following discrete convolution with a mixture of $K$ Gaussian functions
\begin{align}
\phi(t, u) &= \sum_{k=1}^K{\alpha^k_t\expo{-\beta_t^k\left(\kappa_t^k-u\right)^2}}\\
w_t &= \sum_{u=1}^U{\phi(t, u)c_u}
\end{align}
where $\phi(t, u)$ is the \emph{window weight} of $c_u$ at timestep $t$.
Intuitively, the $\kappa_t$ parameters control the location of the window, the $\beta_t$ parameters control the width of the window and the $\alpha_t$ parameters control the importance of the window within the mixture.
The size of the soft window vectors is the same as the size of the character vectors $c_u$ (assuming a one-hot encoding, this will be the number of characters in the alphabet).
Note that the window mixture is not normalised and hence does not determine a probability distribution; however the window weight $\phi(t, u)$ can be loosely interpreted as the network's belief that it is writing character $c_u$ at time $t$.
\fref{window_weights} shows the alignment implied by the window weights during a training sequence.

\fig{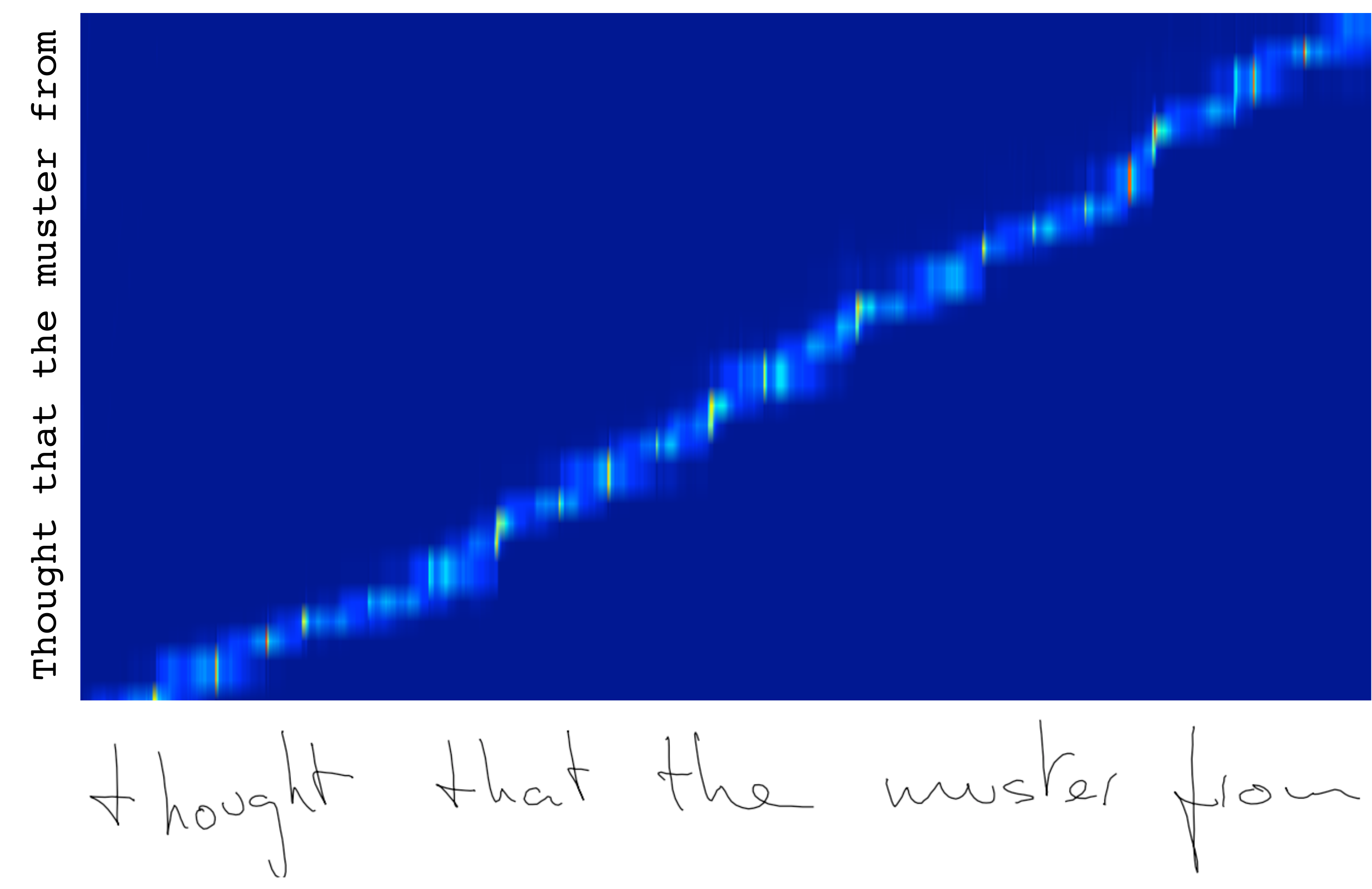}{window_weights}{1}{Window weights during a handwriting synthesis sequence}{ Each point on the map shows the value of $\phi(t, u)$, where $t$ indexes the pen trace along the horizontal axis and $u$ indexes the text character along the vertical axis. The bright line is the alignment chosen by the network between the characters and the writing. Notice that the line spreads out at the boundaries between characters; this means the network receives information about next and previous letters as it makes transitions, which helps guide its predictions.}

The size $3K$ vector $p$ of window parameters is determined as follows by the outputs of the first hidden layer of the network:
\begin{align}
(\hat{\alpha}_t, \hat{\beta}_t, \hat{\kappa}_t) &= W_{h^1 p} h^1_t + b_p\\
{\alpha}_t &= \expo{\hat{\alpha}_t}\\
{\beta}_t &= \expo{\hat{\beta}_t}\\
{\kappa}_t &= \kappa_{t-1} + \expo{\hat{\kappa}_t}
\end{align}
Note that the location parameters ${\kappa}_t$ are defined as offsets from the previous locations $c_{t-1}$, and that the size of the offset is constrained to be greater than zero.
Intuitively, this means that network learns \emph{how far} to slide each window at each step, rather than an absolute location.
Using offsets was essential to getting the network to align the text with the pen trace.

The $w_t$ vectors are passed to the second and third hidden layers at time $t$, and the first hidden layer at time $t+1$ (to avoid creating a cycle in the processing graph).
The update equations for the hidden layers are
\begin{align}
\label{eq:synth_hidden}
h^1_t &= \hiddenfn\left(W_{i h^1} x_t + W_{h^{1}h^{1}} h^1_{t-1} + W_{w h^{1}} w_{t-1} + \hbias^1 \right)\\
h^n_t &= \hiddenfn\left(W_{i h^n} x_t + W_{h^{n-1}h^{n}} h^{n-1}_t + W_{h^{n}h^{n}} h^n_{t-1} + W_{w h^{n}} w_{t} + \hbias^n \right)
\end{align}
The equations for the output layer remain unchanged from Eqs.~(17) to~(22).
The sequence loss is
\begin{equation}
\aloss = -\log \Pr(\inseq|\ann)
\end{equation}
where
\begin{equation}
\Pr(\inseq|\ann) = \prod_{t=1}^{T}{\Pr\left(x_{t+1}|y_t\right)}
\end{equation}
Note that $y_t$ is now a function of $\ann$ as well as $\inseq_{1:t}$.

The loss derivatives with respect to the outputs $\hat{e}_t, \hat{\pi}_t, \hat{\mu}_t, \hat{\sigma}_t, \hat{\rho}_t$ remain unchanged from Eqs.~(27),~(30) and~(31).
Given the loss derivative $\pd{\aloss}{w_t}$ with respect to the size $W$ window vector $w_t$, obtained by backpropagating the output derivatives through the computation graph in \fref{synth_net}, the derivatives with respect to the window parameters are as follows:
\begin{align}
\epsilon(k, t, u) &\deq  {\alpha}^k_t\expo{-\beta_t^k\left(\kappa_t^k-u\right)^2} \sum_{j=1}^{W}{\pd{\aloss}{w^j_t} c^j_u}\\
\pd{\aloss}{\hat{\alpha}^k_t} &= \sum_{u=1}^U{\epsilon(k, t, u)}\\
\pd{\aloss}{\hat{\beta}^k_t} &= -{\beta}^k_t \sum_{u=1}^U{\epsilon(k, t, u) (\kappa^k_t-u)^2}\\
\pd{\aloss}{{\kappa}^k_t} &= \pd{\aloss}{{\kappa}^k_{t+1}} + 2 \beta^k_t \sum_{u=1}^U{\epsilon(k, t, u) (u - \kappa^k_t)}\\
\pd{\aloss}{\hat{\kappa}^k_t} &= \expo{\hat{\kappa}^k_t} \pd{\aloss}{{\kappa}^k_t}
\end{align}

\fref{synth_density} illustrates the operation of a mixture density output layer applied to handwriting synthesis.

\fig{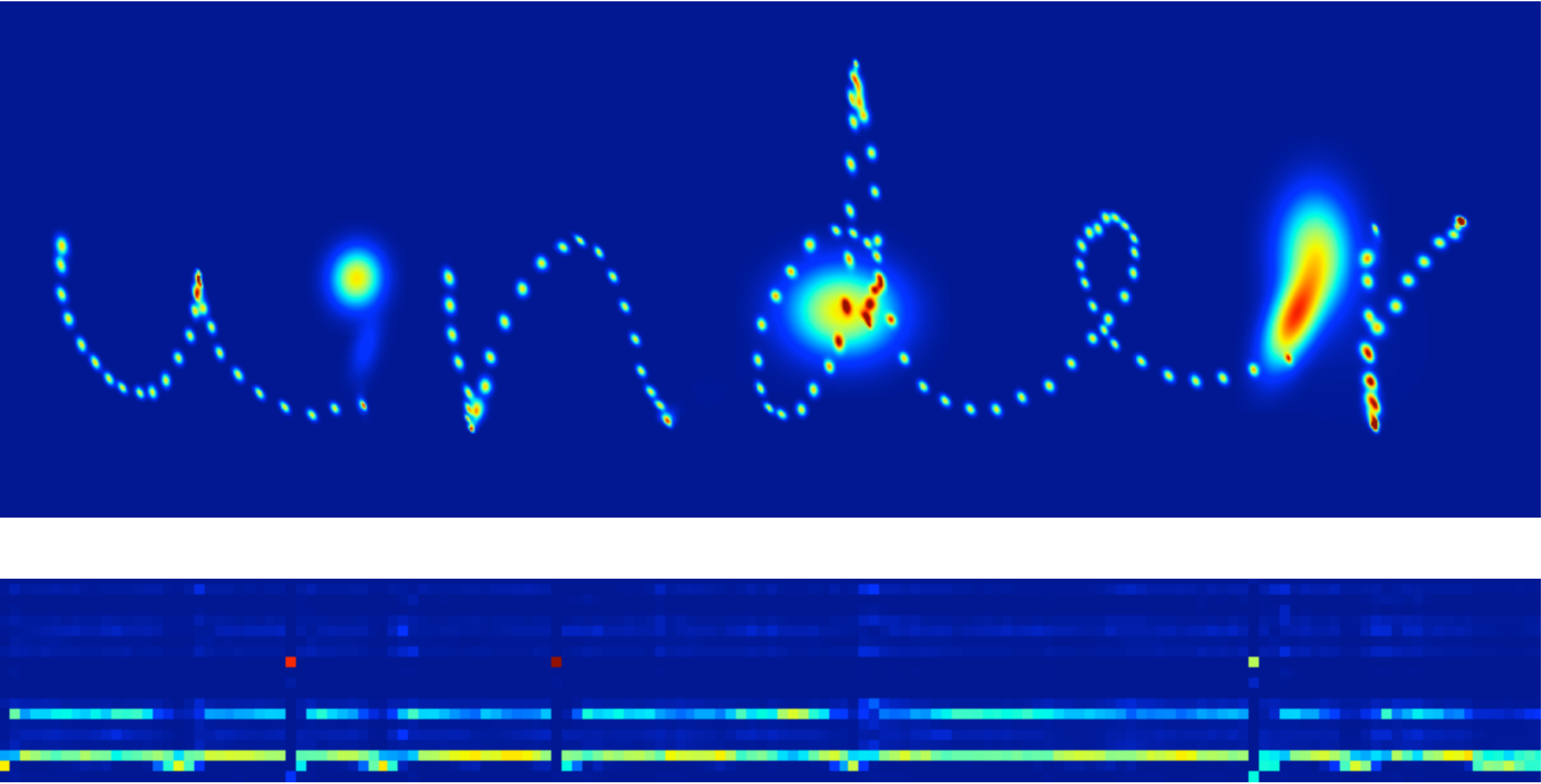}{synth_density}{1}{Mixture density outputs for handwriting synthesis.}{ The top heatmap shows the predictive distributions for the pen locations, the bottom heatmap shows the mixture component weights. Comparison with \fref{pred_density} indicates that the synthesis network makes more precise predictions (with smaller density blobs) than the prediction-only network, especially at the ends of strokes, where the synthesis network has the advantage of knowing which letter comes next.}

\subsection{Experiments}
The synthesis network was applied to the same input data as the handwriting prediction network in the previous section.
The character-level transcriptions from the IAM-OnDB were now used to define the character sequences $\ann$.
The full transcriptions contain 80 distinct characters (capital letters, lower case letters, digits, and punctuation). 
However we used only a subset of 57, with all the digits and most of the punctuation characters replaced with a generic `non-letter' label\footnote{This was an oversight; however it led to the interesting result that when the text contains a non-letter, the network must select a digits or punctuation mark to generate. Sometimes the character can be be inferred from the context (\eg the apostrophe in ``can't''); otherwise it is chosen at random.}.

The network architecture was as similar as possible to the best prediction network: three hidden layers of 400 LSTM cells each, 20 bivariate Gaussian mixture components at the output layer and a size 3 input layer.
The character sequence was encoded with one-hot vectors, and hence the window vectors were size 57.
A mixture of 10 Gaussian functions was used for the window parameters, requiring a size 30 parameter vector.
The total number of weights was increased to approximately 3.7M.

The network was trained with rmsprop, using the same parameters as in the previous section.
The network was retrained with adaptive weight noise, initial standard deviation 0.075, and the output and LSTM gradients were again clipped in the range $[-100,100]$ and $[-10,10]$ respectively.

\tref{hand_synth} shows that adaptive weight noise gave a considerable improvement in log-loss (around 31.3 nats) but no significant change in sum-squared error.
The regularised network appears to generate slightly more realistic sequences, although the difference is hard to discern by eye.
Both networks performed considerably better than the best prediction network.
In particular the sum-squared-error was reduced by 44$\%$.
This is likely due in large part to the improved predictions at the ends of strokes, where the error is largest.

\begin{table}
\centering
\capt{Handwriting Synthesis Results.}{ All results recorded on the validation set. `Log-Loss' is the mean value of $\loss$ in nats. `SSE' is the mean sum-squared-error per data point.}
\tlabel{hand_synth}
\vskip 0.15in
\begin{center}
\begin{sc}
\begin{tabular}{lllll}
\hline
Regularisation &Log-Loss & SSE\\
\hline
none & -1096.9  & 0.23\\
adaptive weight noise & -1128.2 & 0.23\\
\hline
\end{tabular}
\end{sc}
\end{center}
\vskip -0.1in
\end{table}

\subsection{Unbiased Sampling}
Given $\ann$, an unbiased sample can be picked from $\Pr(\inseq|\ann)$ by iteratively drawing $x_{t+1}$ from $\Pr\left(x_{t+1}|y_t\right)$, just as for the prediction network.
The only difference is that we must also decide when the synthesis network has finished writing the text and should stop making any future decisions.
To do this, we use the following heuristic: as soon as $\phi(t, U+1) > \phi(t, u)\ \forall\ 1 \leq u \leq U$ the current input $x_t$ is defined as the end of the sequence and sampling ends.
Examples of unbiased synthesis samples are shown in \fref{synth_val}.
These and all subsequent figures were generated using the synthesis network retrained with adaptive weight noise. 
Notice how stylistic traits, such as character size, slant, cursiveness etc. vary widely between the samples, but remain more-or-less consistent within them.
This suggests that the network identifies the traits early on in the sequence, then remembers them until the end.
By looking through enough samples for a given text, it appears to be possible to find virtually any combination of stylistic traits, which suggests that the network models them independently both from each other and from the text.

`Blind taste tests' carried out by the author during presentations suggest that at least some unbiased samples cannot be distinguished from real handwriting by the human eye. 
Nonetheless the network does make mistakes we would not expect a human writer to make, often involving missing, confused or garbled letters\footnote{We expect humans to make mistakes like misspelling `temperament' as `temperement', as the second writer in \fref{synth_val} seems to have done.}; this suggests that the network sometimes has trouble determining the alignment between the characters and the trace.
The number of mistakes increases markedly when less common words or phrases are included in the character sequence.
Presumably this is because the network learns an implicit character-level language model from the training set that gets confused when rare or unknown transitions occur.

\begin{figure}
\includegraphics[width=0.95\columnwidth]{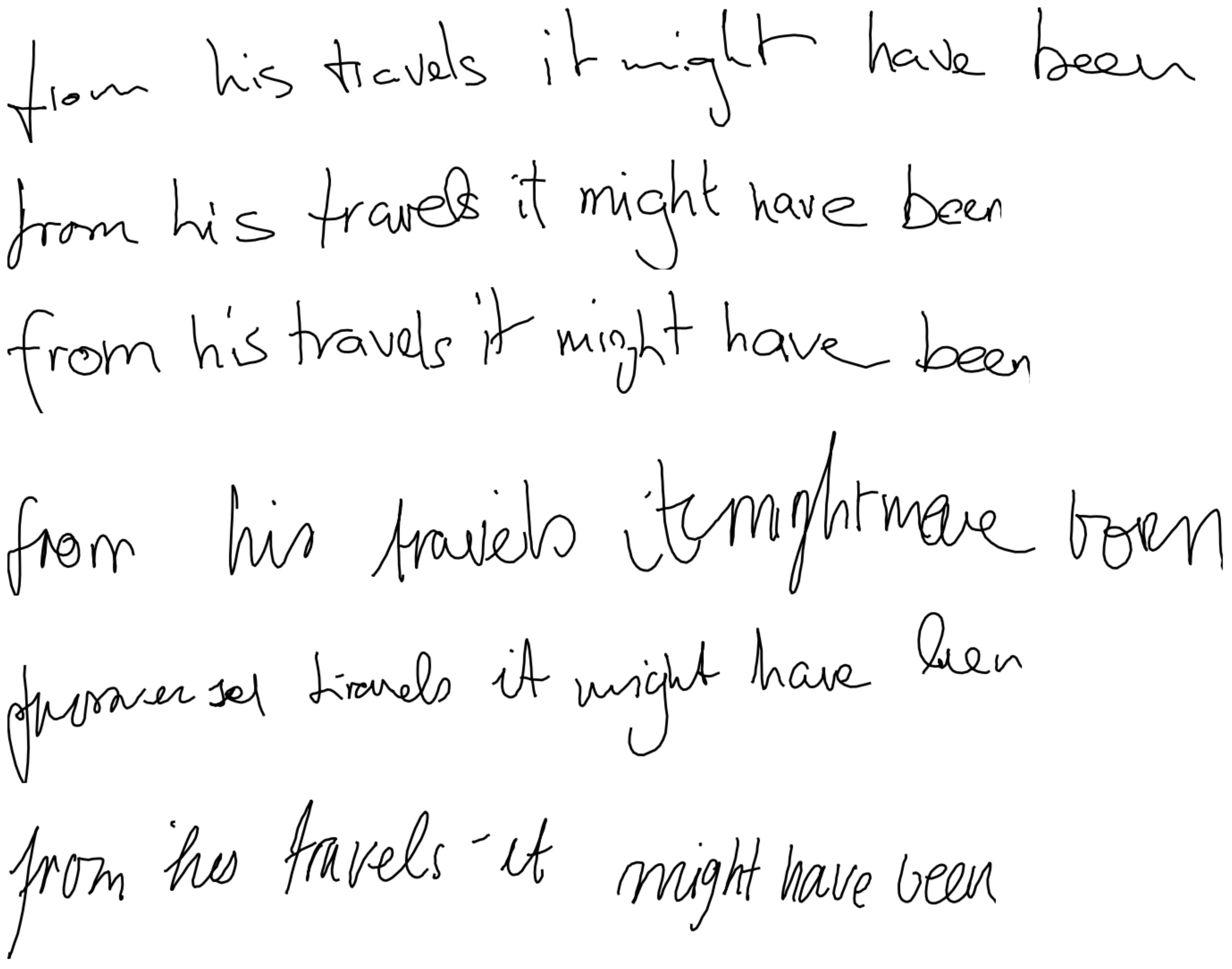}\\
\\
\\
\\
\includegraphics[width=0.95\columnwidth]{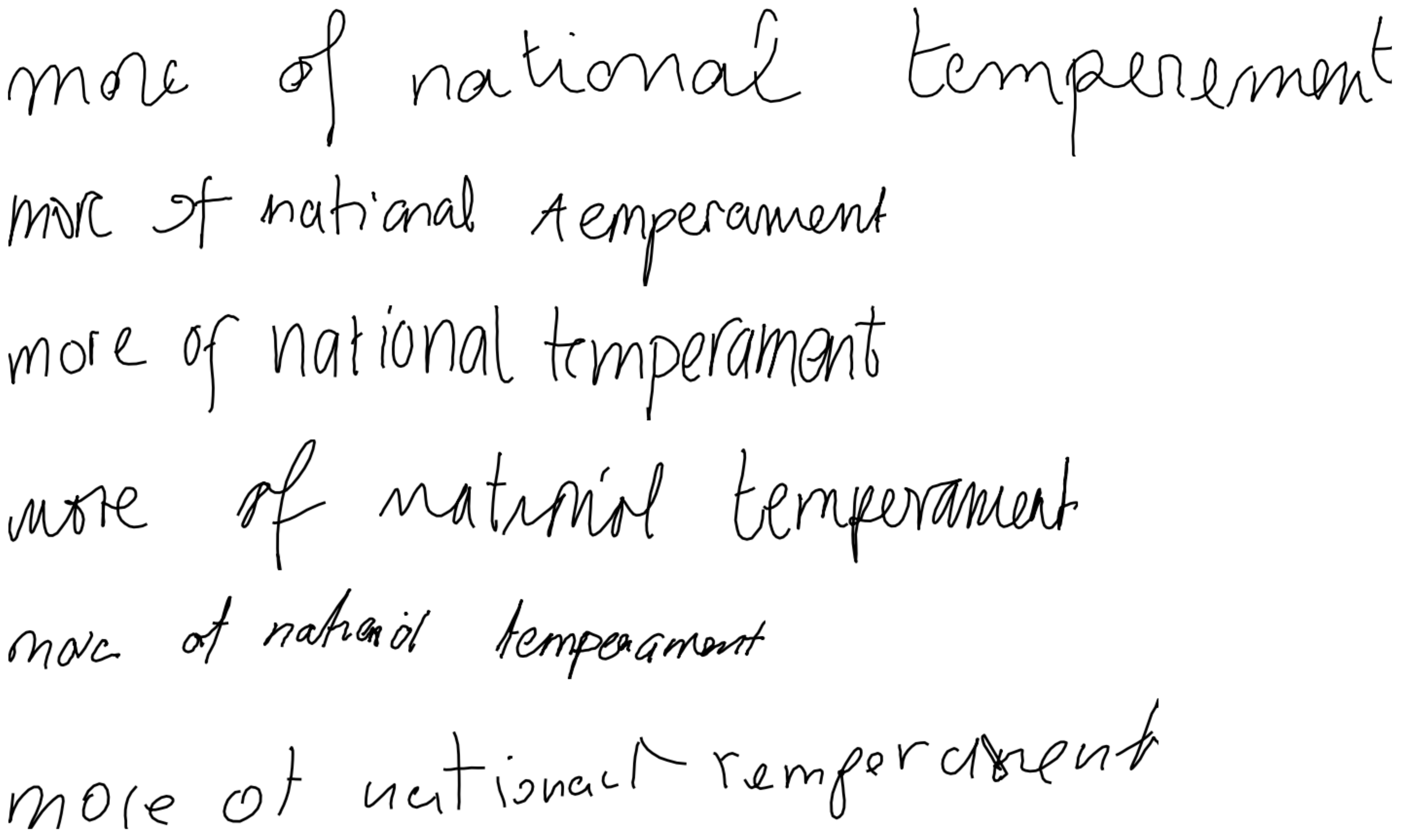}
\\
\caption{\textbf{Real and generated handwriting}. The top line in each block is real, the rest are unbiased samples from the synthesis network. The two texts are from the validation set and were not seen during training.}
\flabel{synth_val}
\end{figure}


\subsection{Biased Sampling}
One problem with unbiased samples is that they tend to be difficult to read (partly because real handwriting is difficult to read, and partly because the network is an imperfect model).
Intuitively, we would expect the network to give higher probability to good handwriting because it tends to be smoother and more predictable than bad handwriting.
If this is true, we should aim to output more probable elements of $\Pr(\inseq|\ann)$ if we want the samples to be easier to read. 
A principled search for high probability samples could lead to a difficult inference problem, as the probability of every output depends on all previous outputs.
However a simple heuristic, where the sampler is biased towards more probable predictions at each step independently, generally gives good results.
Define the \emph{probability bias} $b$ as a real number greater than or equal to zero.
Before drawing a sample from $\Pr(x_{t+1}|y_t)$, each standard deviation $\sigma^j_t$ in the Gaussian mixture is recalculated from Eq.~(21) to 
\begin{equation}
\sigma^j_t = \expo{\hat{\sigma}^j_t - b}
\end{equation}
and each mixture weight is recalculated from Eq.~(19) to
\begin{equation}
\pi^j_t = \frac{\expo{\hat{\pi}^j_t(1+b)}}{\sum_{j'=1}^M{\expo{\hat{\pi}^{j'}_t(1+b)}}}
\end{equation}
This artificially reduces the variance in both the choice of component from the mixture, and in the distribution of the component itself.
When $b=0$ unbiased sampling is recovered, and as $b \rightarrow \infty$ the variance in the sampling disappears and the network always outputs the mode of the most probable component in the mixture (which is not necessarily the mode of the mixture, but at least a reasonable approximation).
\fref{synth_biased} shows the effect of progressively increasing the bias, and \fref{synth_val_bias} shows samples generated with a low bias for the same texts as \fref{synth_val}.

\fig{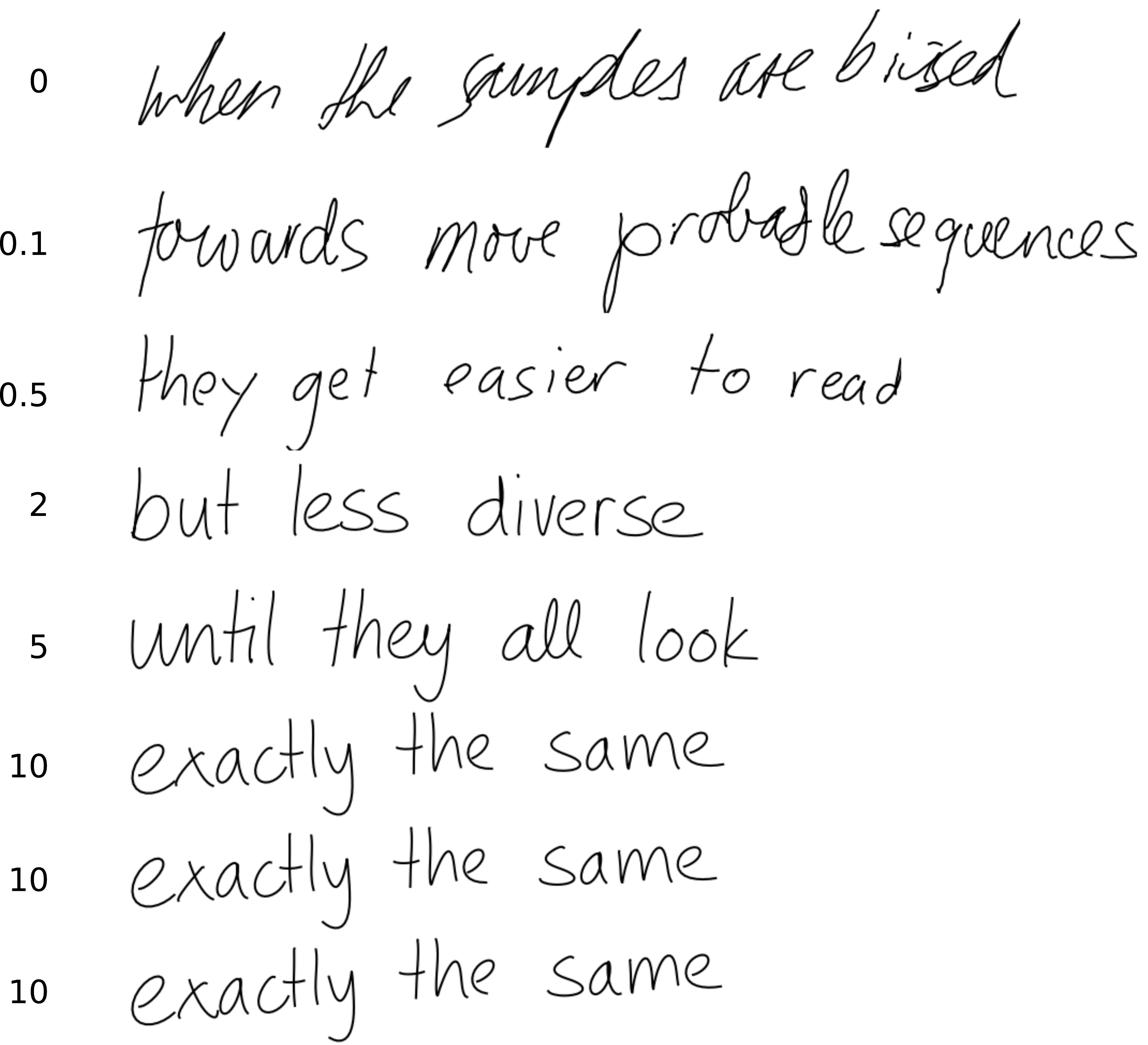}{synth_biased}{0.9}{Samples biased towards higher probability.}{ The probability biases $b$ are shown at the left. As the bias increases the diversity decreases and the samples tend towards a kind of `average handwriting' which is extremely regular and easy to read (easier, in fact, than most of the real handwriting in the training set).
Note that even when the variance disappears, the same letter is not written the same way at different points in a sequence (for examples the `e's in ``exactly the same'', the `l's in ``until they all look''), because the predictions are still influenced by the previous outputs.
If you look closely you can see that the last three lines are not quite exactly the same.}

\begin{figure}
\includegraphics[width=0.95\columnwidth]{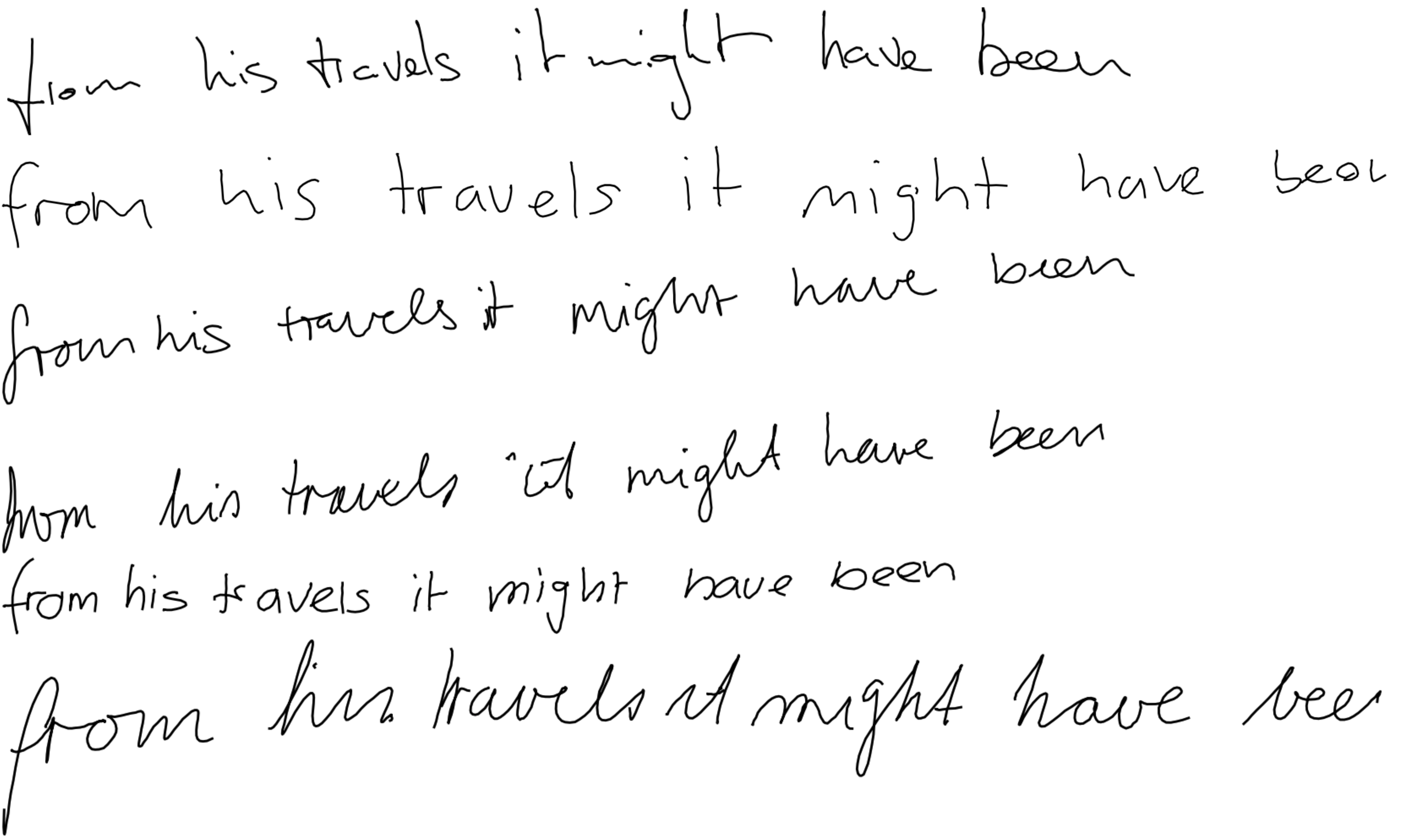}\\
\\
\\
\\
\includegraphics[width=0.95\columnwidth]{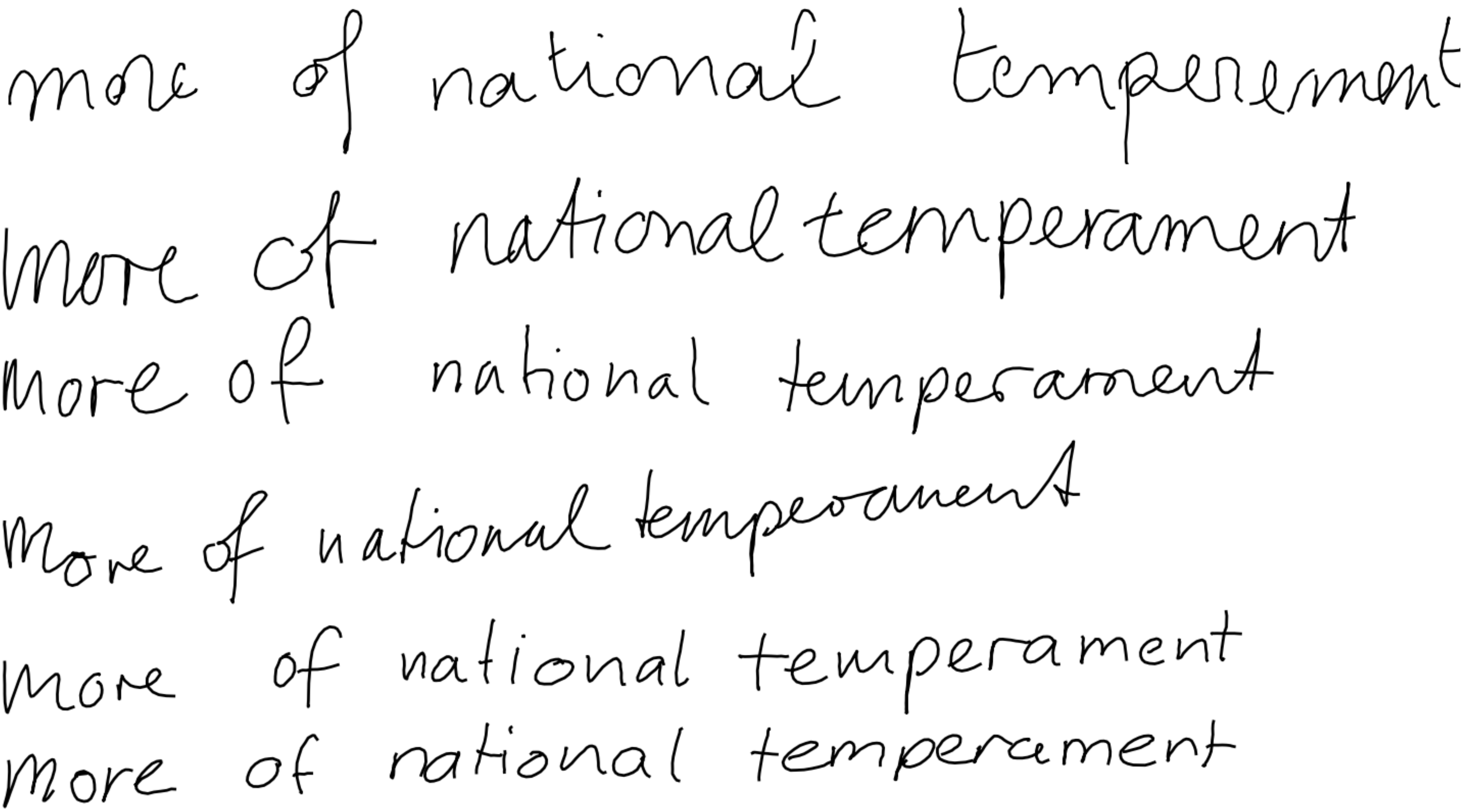}
\\
\caption{\textbf{A slight bias.} The top line in each block is real. The rest are samples from the synthesis network with a probability bias of 0.15, which seems to give a good balance between diversity and legibility.}
\flabel{synth_val_bias}
\end{figure}

\subsection{Primed Sampling}
Another reason to constrain the sampling would be to generate handwriting in the style of a particular writer (rather than in a randomly selected style).
The easiest way to do this would be to retrain it on that writer only.
But even without retraining, it is possible to mimic a particular style by `priming' the network with a real sequence, then generating an extension with the real sequence still in the network's memory.
This can be achieved for a real $\inseq$, $\ann$ and a synthesis character string $\seq{s}$ by setting the character sequence to $\ann'=\ann+\seq{s}$ and clamping the data inputs to $\inseq$ for the first $T$ timesteps, then sampling as usual until the sequence ends.
Examples of primed samples are shown in Figs.~\ref{fig:primed_samples_1} and~\ref{fig:primed_samples_2}.
The fact that priming works proves that the network is able to remember stylistic features identified earlier on in the sequence.
This technique appears to work better for sequences in the training data than those the network has never seen.

\fig{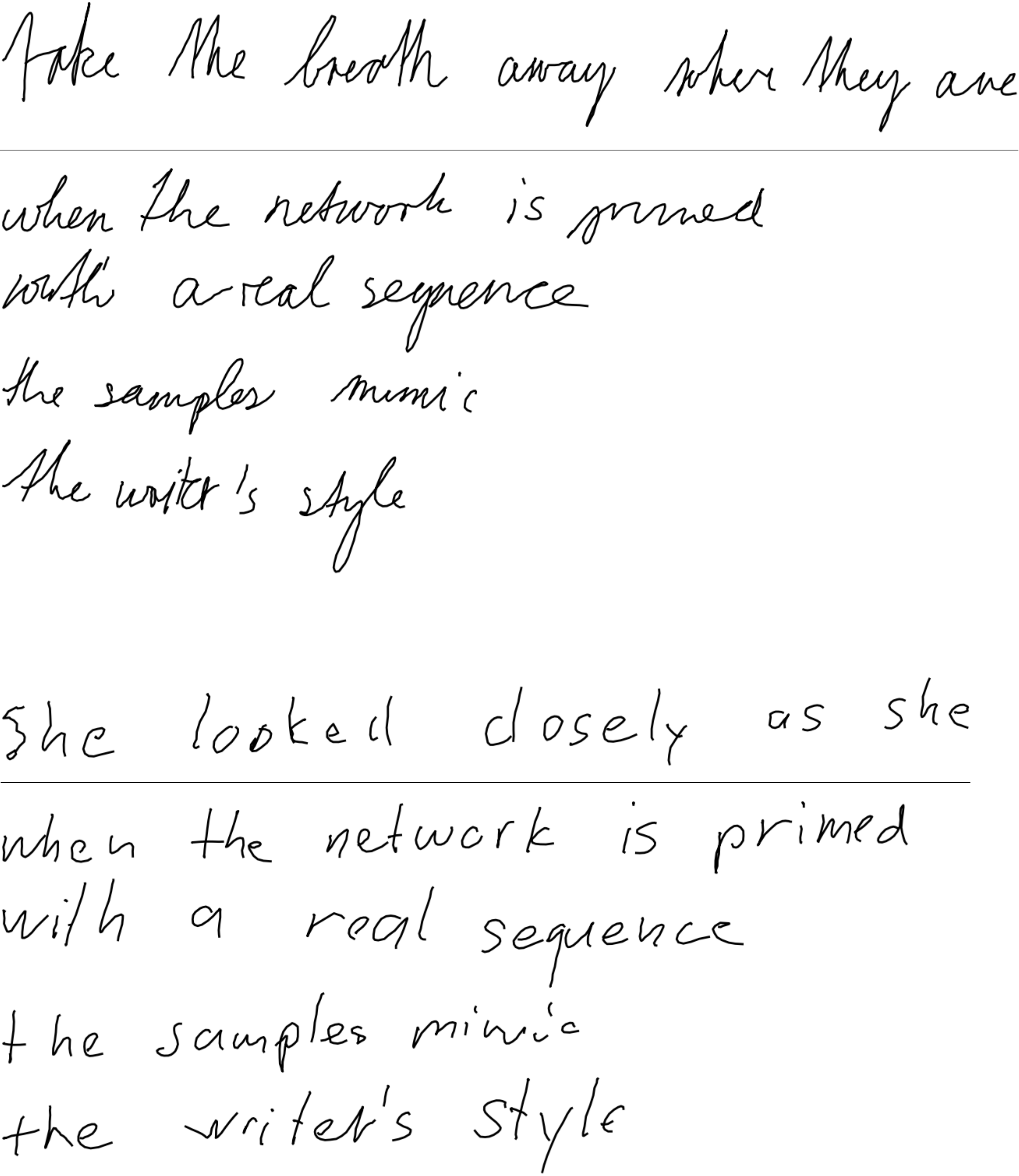}{primed_samples_1}{1}{Samples primed with real sequences.}{ The priming sequences  (drawn from the training set) are shown at the top of each block. None of the lines in the sampled text exist in the training set. 
The samples were selected for legibility.}
\fig{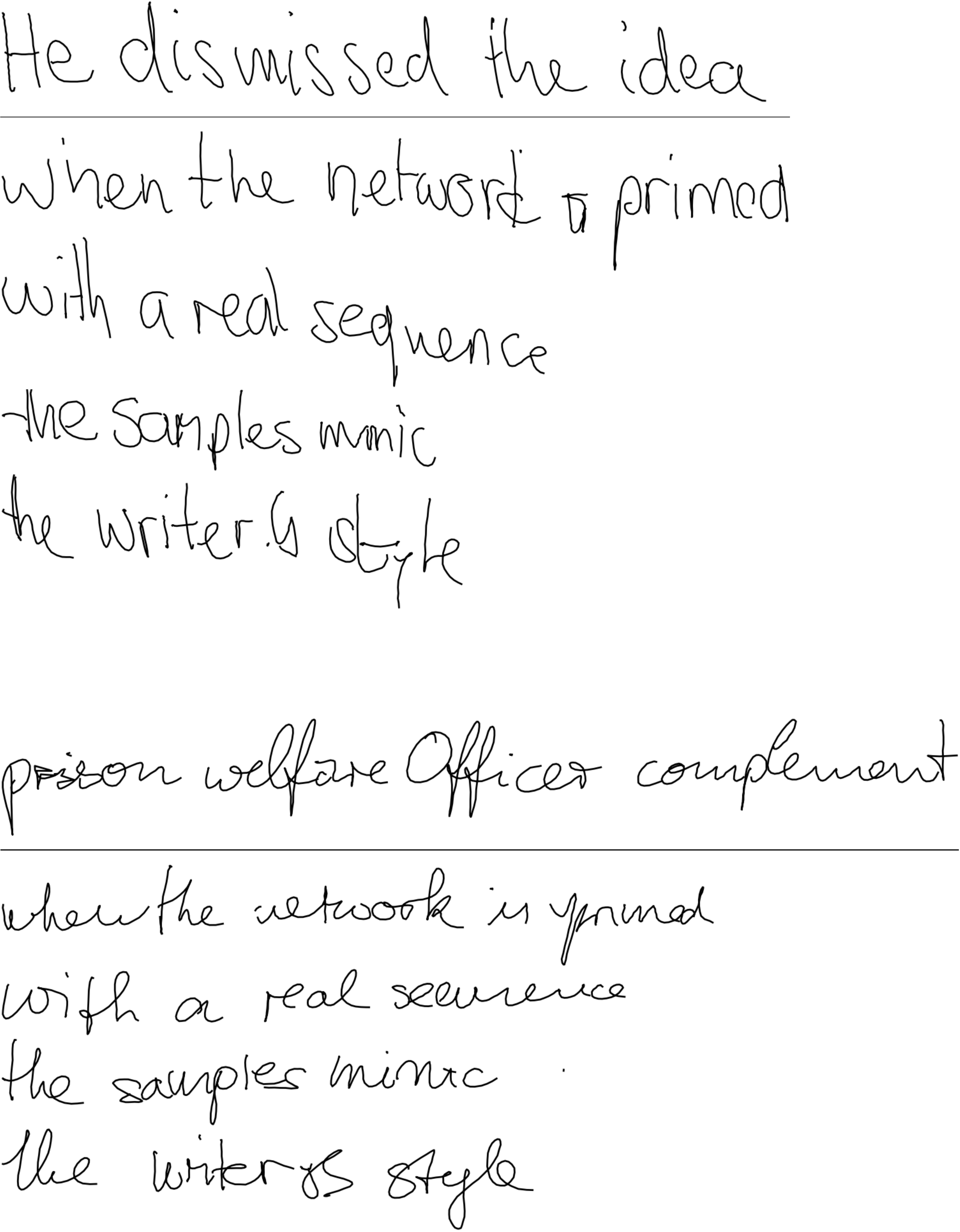}{primed_samples_2}{1}{Samples primed with real sequences (cotd).}

Primed sampling and reduced variance sampling can also be combined.
As shown in Figs.~\ref{fig:bias_primed_1} and~\ref{fig:bias_primed_2} this tends to produce samples in a `cleaned up' version of the priming style, with overall stylistic traits such as slant and cursiveness retained, but the strokes appearing smoother and more regular.
A possible application would be the artificial enhancement of poor handwriting.

\fig{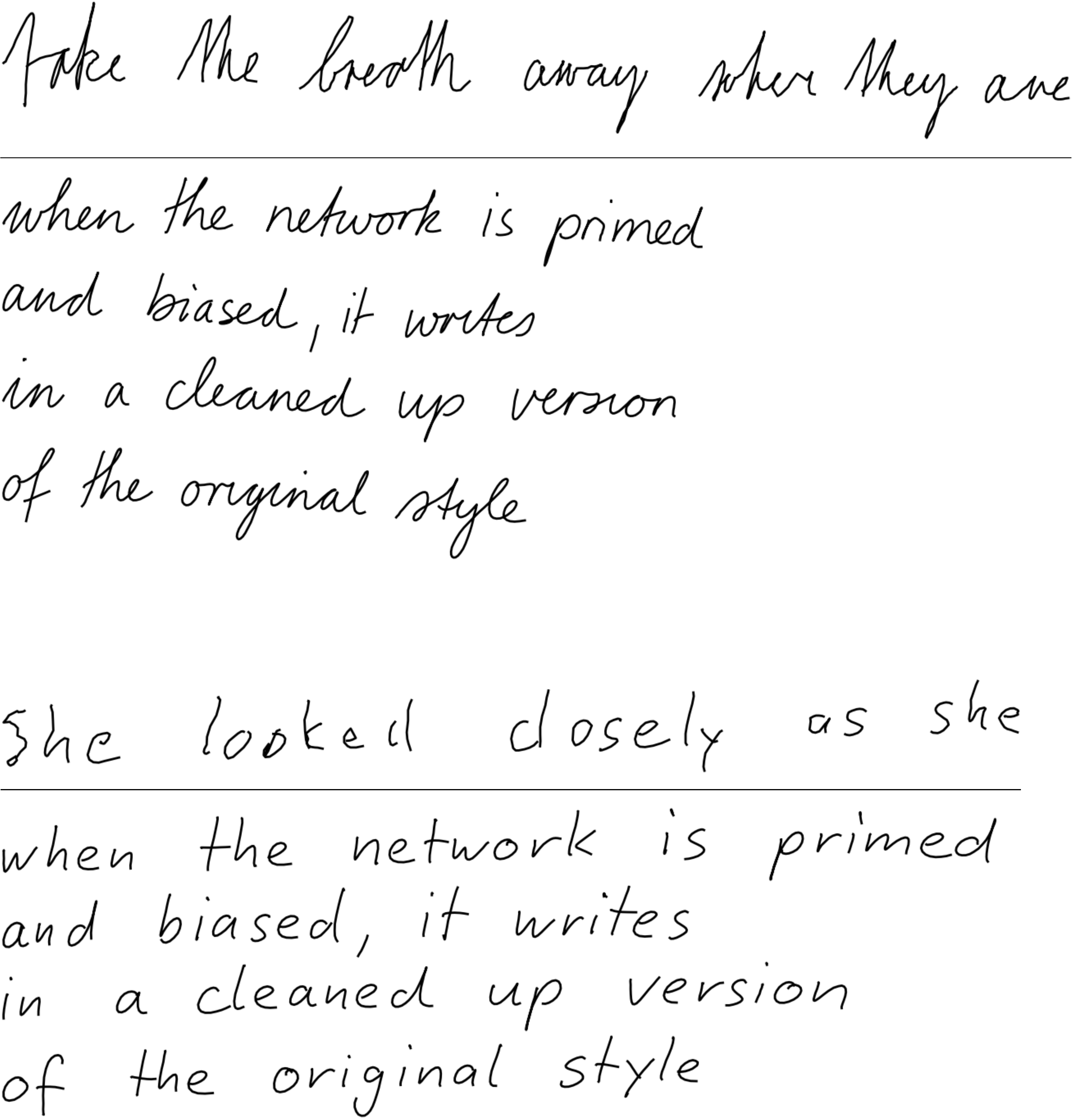}{bias_primed_1}{1}{Samples primed with real sequences \emph{and} biased towards higher probability.}{ The priming sequences are at the top of the blocks. The probability bias was 1. None of the lines in the sampled text exist in the training set.}

\fig{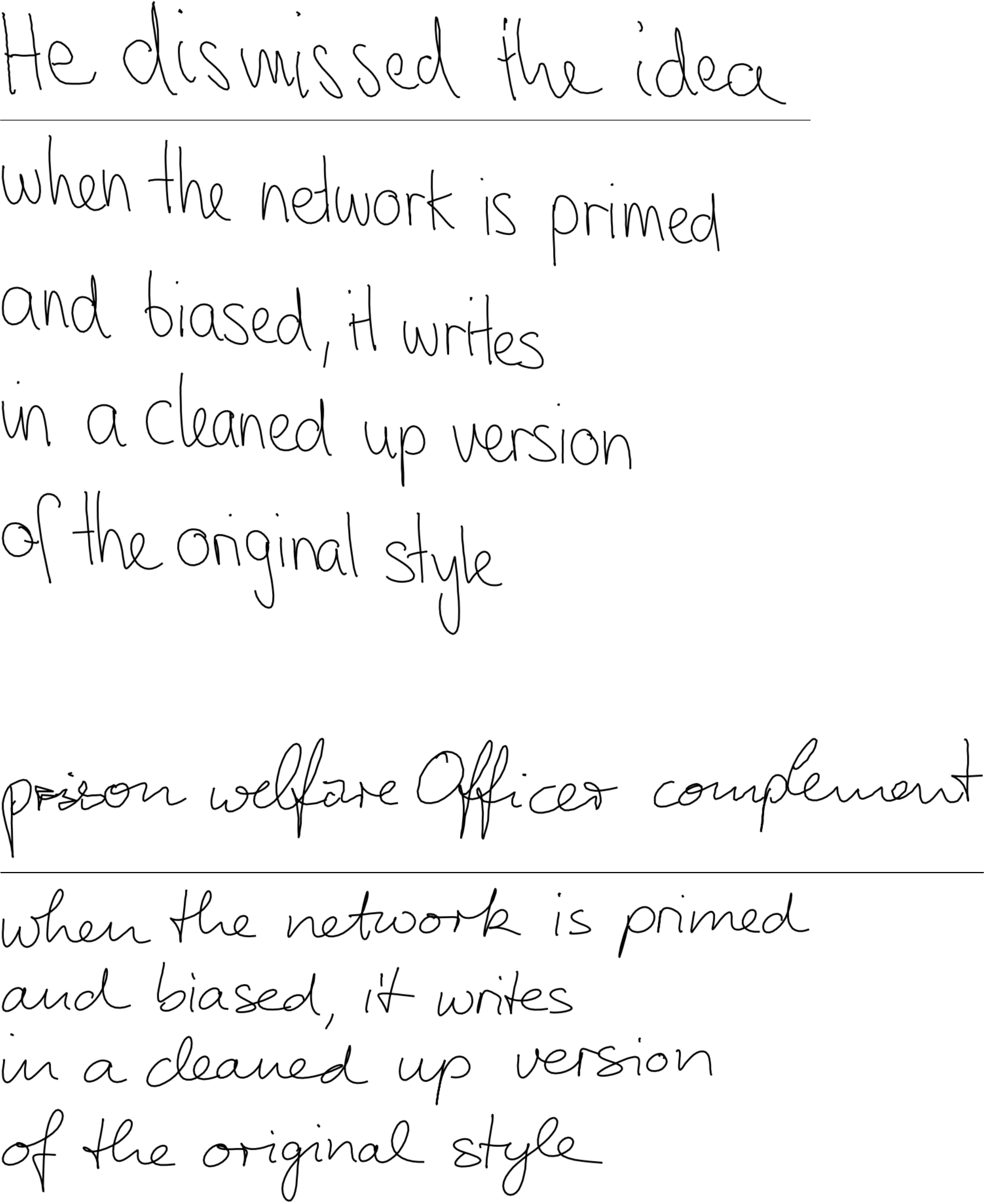}{bias_primed_2}{1}{Samples primed with real sequences \emph{and} biased towards higher probability (cotd)}


\section{Conclusions and Future Work}
\seclabel{conclusion}
This paper has demonstrated the ability of Long Short-Term Memory recurrent neural networks to generate both discrete and real-valued sequences with complex, long-range structure using next-step prediction.
It has also introduced a novel convolutional mechanism that allows a recurrent network to condition its predictions on an auxiliary annotation sequence, and used this approach to synthesise diverse and realistic samples of online handwriting.
Furthermore, it has shown how these samples can be biased towards greater legibility, and how they can be modelled on the style of a particular writer.

Several directions for future work suggest themselves.
One is the application of the network to speech synthesis, which is likely to be more challenging than handwriting synthesis due to the greater dimensionality of the data points.
Another is to gain a better insight into the internal representation of the data, and to use this to manipulate the sample distribution directly.
It would also be interesting to develop a mechanism to automatically extract high-level annotations from sequence data.
In the case of handwriting, this could allow for more nuanced annotations than just text, for example stylistic features, different forms of the same letter, information about stroke order and so on.

\section*{Acknowledgements}
Thanks to Yichuan Tang, Ilya Sutskever, Navdeep Jaitly, Geoffrey Hinton and other colleagues at the University of Toronto for numerous useful comments and suggestions. 
This work was supported by a Global Scholarship from the Canadian Institute for Advanced Research.

\bibliographystyle{abbrv}
{
\bibliography{rnn_sequences}
}

\end{document}